\documentclass[runningheads]{llncs}
\usepackage[T1]{fontenc}
\usepackage{graphicx}
\usepackage{booktabs}
\usepackage[misc]{ifsym}
\newcommand{\corr}{(\Letter)}
\usepackage{amsmath}
\usepackage{mwe}
\usepackage{hyperref}
\usepackage{siunitx}
\usepackage{subcaption}
\begin{document}
\title{Evaluating Vision Transformer Models for Visual Quality Control in Industrial Manufacturing}
\titlerunning{ViT Models for Industrial Visual Quality Control}
%N.B.: Author information (both in the \author{} and \authorrunning{} command) should only be present in the Camera-Ready Version of your paper. The version that you initially submit for review, ought to be double-blind. So, when initially submitting your paper, use:
\author{Author information scrubbed for double-blind reviewing}
% exclude author / institution information for anonymity
\author{Miriam  Alber\inst{1,3} \corr\and
Christoph H{\"o}nes\inst{2} \and
Patrick Baier\inst{3}}
%
% You may leave out the orcidID information, if you want to.
% Use \corr to indicate the corresponding author. Note the spacing around the \corr command. Only one author can be the corresponding author.
\authorrunning{M. Alber et al.}
% exclude author / institution information for anonymity
\institute{iteratec GmbH Karlsruhe, Germany \email{miriam.alber@iteratec.com} \and
Hasso Plattner Institute / University of Potsdam, Germany
\and
University of Applied Sciences Karlsruhe, Germany}
\toctitle{Evaluating Vision Transformer Models for Visual Quality Control in Industrial Manufacturing}
\tocauthor{Miriam~Alber,Christoph~H{\"o}nes,Patrick~Baier}
\maketitle              % typeset the header of the contribution
\begin{abstract}
% The abstract should briefly summarize the contents of the paper in
% 15--250 words.
One of the most promising use-cases for machine learning in industrial manufacturing is the early detection of defective products using a quality control system. Such a system can save costs and reduces human errors due to the monotonous nature of visual inspections.
Today, a rich body of research exists which employs machine learning methods to identify rare defective products in unbalanced visual quality control datasets. 
These methods typically rely on two components: A visual backbone to capture the features of the input image and an anomaly detection algorithm that decides if these features are within an expected distribution. With the rise of transformer architecture as visual backbones of choice, there exists now a great variety of different combinations of these two components, ranging all along the trade-off between detection quality and inference time.
Facing this variety, practitioners in the field often have to spend a considerable amount of time on researching the right combination for their use-case at hand. Our contribution is to help practitioners with this choice by reviewing and evaluating current vision transformer models together with anomaly detection methods. For this, we chose SotA models of both disciplines, combined them and evaluated them towards the goal of having small, fast and efficient anomaly detection models suitable for industrial manufacturing. We evaluated the results of our experiments on the well-known MVTecAD and BTAD datasets. Moreover, we give guidelines for choosing a suitable model architecture for a quality control system in practice, considering given use-case and hardware constraints.

\end{abstract}

\section{Introduction}
In industrial manufacturing, early detection of defective products saves material and costs and enhances public trust in the manufacturer. Automating this process increases scalability, saves labour costs and reduces human error due to the monotonous nature of visual inspections~\cite{Wang.2018}. To this end, the possibility of automating this process using machine learning methods has been subject of extensive research~\cite{Tao.2022}.
Anomaly detection (AD) in machine learning addresses the challenge of identifying rare defective products in unbalanced datasets, often utilizing unsupervised or semi-supervised training strategies. While AD refers to image-level classification of samples as either normal or anomalous, anomaly localization (AL) aims to identify anomalies on a more fine-grained level and indicates where the anomalous feature was detected within the image.
This provides interpretability of the model's decisions, facilitating human-in-the-loop control by allowing focus on the anomalous regions.
Unsupervised AD architectures typically consist of an image encoding backbone and a detection algorithm that identifies if the extracted features are within an expected distribution.
Since the publication of \textit{ViT} in 2020~\cite{Dosovitskiy.2020}, vision transformers have emerged as an alternative to traditional CNN backbones, offering enhanced global dependency capture, making them particularly interesting for AD tasks. Emerging hierarchical transformer architectures promise to solve the problem of their excessive size by keeping their advantages~\cite{Vaswani.2021}.
Many approaches to solve AD problems in existing literature consider monolithic vision transformer but not hierarchical ones \cite{Mishra.2021,Yu.2021}. We believe hierarchical vision transformer models can be a great benefit for industrial visual quality control regarding their computational and memory demands. Moreover, we want to help practitioners with their choice of the best setup by reviewing and evaluating current vision transformer models together with AD methods. Our contributions can be summarized as follows:
\begin{enumerate}
    \item First, we provide a comprehensive overview of current state-of-the-art (SotA) hierarchical vision transformer models and approaches for the task of visual quality control.
    \item Second, we reproduce two of the most promising AD methods and combine them with different visual backbones to find small, fast and efficient AD models, suitable for industrial manufacturing, and evaluate our results on the well-known MVTecAD and BTAD datasets.
    \item At last, we recommend guidelines for choosing a suitable architecture for deploying a quality control system in practice considering given use-case and hardware constraints.
\end{enumerate}
The rest of this paper is structured as follows: In the next section, we  review existing work regarding visual backbones, AD and AL. 
Section~\ref{sec:experiments} describes the setup we chose for our experiments, which are then discussed and analysed in Section
~\ref{sec:results}. Our source code is available on GitHub (\url{https://github.com/visiontransformerad/vit-ad}).

\section{Related Work}
% \section{Visual Quality Control}
%Before we start describing our experiments, we shortly give an overview of current state-of-the-art techniques in the field of visual quality control. As described before, such systems consist of a vision backbone and, on top of it, a system for AD and AL.

\subsection{Vision Backbone}
Vision transformers have become a powerful alternative to CNNs for various computer vision tasks, with many studies highlighting their proficiency in capturing global dependencies, which are essential for AD and AL workloads~\cite{Mathian.862022,You.952022,Choi.2022,Mishra.2021,Yu.2021}.
% \cite{Mathian.862022,You.952022,Choi.2022,Mishra.2021,Pirnay.4282021,YOASEUNGDONGLEESEUNGJUNKIMCHIYOONJ.KHYUNWOO.1132023}.
% %previous text
% \iffalse
% The automation of visual quality control with image-level AD has been researched for over a decade \cite{Pimentel.2014}.
% Nevertheless, progress in new CV technologies offers opportunities to continually improve existing work
% in this field. One example is the adaption of transformer models for CV tasks \cite{Dosovitskiy.2020}. Several works have
% been published, which combine CNNs with vision transformer models or even replace them entirely \cite{Khan.2022,Jamil.11112022}. First, we outline current research in the field of pure vision transformers as a replacement for CNN backbones. Then we will present current research in the field of visual anomaly detection and localization.
% \fi
% copied from thesis 3.1
Monolithical vision transformers such as \textit{ViT} follow the architecture of NLP transformers, offering performance competitive with CNNs like \textit{ResNet}~\cite{He.2015ResNet} but demand more training data, memory, and longer inference times. By applying an enhanced training strategy using a teacher model and a distillation token to match it's output, Touvron et al. \cite{Touvron.2020} published \textit{DeiT}, an improved monolithic transformer that is smaller than \textit{ViT}, demands less training data and has a superior performance.
% The first competitive approach in using transformers for computer vision tasks was the vision transformer \textit{ViT} \cite{Dosovitskiy.2020}, using nearly the same architecture as for NLP tasks \cite{Vaswani.AttentionIsAll}. These monolithic vision transformers offer performance competitive with CNNs like \textit{ResNet} \cite{He.2015ResNet} but demand more training data, memory, and longer inference times. 
% Touvron et al. \cite{Touvron.2020} introduce their \textit{DeiT} Transformer that removes the dependency on large datasets by using knowledge distillation, i.e. they add a distillation token that is trained to match the output of a teacher model. 
% Chen et al. introduce \textit{CrossViT}, adding a branch for multi-scale patch processing to transformers, enhancing fine-grained analysis with minimal efficiency loss \cite{Chen.2021c}.
However, in light of the limited computational resources in production settings, the light-weight class of hierarchical vision transformers gained interest in recent work~\cite{Vaswani.2021,Wang.2022,Liu.2021,Li.6172021,Zhang.5262021}.
% \cite{Vaswani.2021,Wang.2022,Liu.2021,Li.6172021,Zhang.5262021,Lee.2022MPViT}.
The main idea is to address the poor scaling of monolithic transformers to high-resolution images by reducing the image size across layers.
This leads to models that need less time and resources for training and inference but perform equal to existing vision transformer models~\cite{Vaswani.2021}.
% DeitIII is a follow up, deleted for reasons of space \cite{Touvron2022DeiTIR}
% \iffalse
% The transformer models presented so far mostly follow the architecture in \cite{Vaswani.AttentionIsAll} and do not perform any
% compression of the original image. Since this scales badly when it comes to high-resolution images,
% further research evaluated approaches to bring the idea of CNNs into pure transformer architecture \cite{Graham.422021,Vaswani.2021,wang2021pyramid,Liu.2021,Li.6172021,Zhang.5262021,Lee.2022MPViT}.
% \fi
\textit{HaloNet}~\cite{Vaswani.2021} is one of the first approaches of using size-reducing layers in combination with attention in the encoder, inspired by the architecture used in \textit{ResNet}~\cite{He.2015ResNet}. Wang et al. introduced \textit{PVT}, a model with an attention backbone suitable not only for classification but also for dense prediction tasks such as object detection and segmentation~\cite{Wang.2022}. Zhang et al.~\cite{Zhang.5262021} propose with \textit{NesT} a lightweight yet not overly complex architecture by applying the \textit{ViT} architecture~\cite{Dosovitskiy.2020} on distinct subsets (blocks) of patches and subsequently reducing groups of four neighbouring blocks into one.
Liu et al. developed a similar approach to \textit{PVT} but highlight, that their model has linear instead of quadratic complexity when it comes to scaling with image size~\cite{Liu.2021}. Their \textit{SwinTransformer} model alternates between a window-based and shifted window-based self-attention, which computes attention locally on a fixed number of patches within non-overlapping windows. To enable cross-window connections, non-overlapping neighboring windows from the previous layer are included in the calculations of each block. For dimension reduction, a patch merging layer is added before each stage, which itself can consist of two or more transformer blocks.
% % previous paragraph
% \iffalse
% \paragraph{Swin Transformer}
% Liu et al. developed a similar approach to PVT but highlight, that their model has linear instead of quadratic complexity when it comes to scaling with image size~\cite{Liu.2021}. Their SwinTransformer model alternates between a window-based and shifted window-based SA, which computes attention locally on a fixed
% number of patches with non-overlapping windows. 
% Similar to Wang et al.~\cite{wang2021pyramid}, the original transformer block from~\cite{Dosovitskiy.2020} is used and multi-head SA is replaced with their custom implementation. 
% The model consists of several stages with two or more transformer blocks per stage. 
% For dimension reduction, a patch merging layer is added before each stage. To enable cross-window connections, non-overlapping neighboring windows from the previous layer are included in the calculations of each block.
% Furthermore, the keys are shared between all query patches inside one window. As a result, the Swin Transformer model is presented as a versatile backbone for image classification and dense prediction tasks, which scales linearly with the image size.
% \fi
Li et al.~\cite{Li.6172021} highlight the issue of multi-stage vision transformers with sparse self-attention failing to detect fine-grained inter-region dependencies. They introduce a label-free knowledge distillation strategy, using altered image views for training. Their approach, dubbed \textit{EsViT}, combines view and region-level prediction losses. Their training procedure can be applied to various transformer architectures and pre-trained weights are available for~\cite{Liu.2021,Wang.2022} and~\cite{Touvron.2020}.
Li et al.~\cite{li2022efficientformer} introduced the \textit{EfficientFormer}, an approach that is comparable in inference speed with lightweight CNN implementations such as \textit{MobileNetV2}~\cite{Sandler.} and thus can be used in edge applications. To achieve this, they observed the main bottlenecks (e.g linear projection layers for patch embeddings, reshape operations) in the existing vision transformers and tried to improve efficiency with a few architecture modifications.

\subsection{Anomaly Detection and Localization}
\label{sec:ad_representation_based}
There are several different categories of approaches for implementing AD and AL in practice~\cite{Tao.2022}.
Reconstruction-based methods use an auto-encoder with the objective to compress the input to a lower dimensional latent space and subsequently reconstruct the original image, using the reconstruction error as anomaly score. They tend to deliver weaker results than other categories~\cite{You.952022}. Self-supervised learning methods try to use synthetic data to train a model, which requires costly data augmentation strategies that often heavily depend on expert knowledge. % we want to design a simple and reusable approach. Generating synthetic data needs knowledge about each kind of anomaly which we may not have
In contrast, promising representation-based approaches try to model the distribution of normal features and classify defects as samples that are in regions with low probability density or outside of the modeled distribution~\cite{Tao.2022}.
Roth et al.~\cite{Roth.6152021} achieved excellent detection scores with \textit{PatchCore} by using a memory bank and measuring the distance of extracted features to the nearest feature in the bank. Bae et al.~\cite{Bae.11222022} enhance \textit{PatchCore} with position and neighborhood information to improve the detection of global anomalies. Hyun et al. introduce \textit{ReConPatch}, that uses multiple levels of a CNN encoder as base to create patch-level features, also enhanced by neighborhood information~\cite{Hyun.5262023}. Li et al. introduce \textit{SemiREST}, a model that uses supervised and semi-supervised learning in combination with the \textit{SwinTransformer} architecture~\cite{Li.2023}.
% Within the family of representation-based methods, two approaches stand out for their impressive performance \cite{Tao.2022} and will be discussed in more detail:

Gaussian Mixture Models (GMM) are based on the assumption, that the underlying distribution of normal features is more complex than a single Gaussian distribution and thus learns a set of distributions~\cite{Bishop.}. At inference time, a GMM estimates the probability that a feature belongs to the set of learned distributions~\cite{Mishra.2021}.
Zong et al.~\cite{Zong.} applied an auto-encoder model for AD on non-image data, integrating a GMM for both dense prediction and regularization to avoid local minima in training.
% Zong et al.~\cite{Zong.} used this approach in 2018 to perform AD with an AE model on non-image data. They jointly trained the GMM for dense prediction with the created embedding and the reconstruction loss. The GMM serves as a regularization for the AE and helps not to get stuck in local minima. % As a result, the encoder can create an improved embedding for density estimation without pretraining. %
Zhang et al.~\cite{Zhang.2022} developed a three-component model for AD in high-resolution images, featuring class-specific training and evaluation. This model includes a patch embedding creator, a GMM for density prediction, and a multi-layer perceptron (MLP) for location prediction, jointly trained end-to-end. This approach yields a smaller model compared to pre-trained alternatives.
% Zhang et al. adopted their approach for AD in high-resolution image data and propose a three-component model that is trained and evaluated class-specifically~\cite{Zhang.2022}. The first component creates a low-dimensional patch embedding, which is then used by the second component, a GMM, for densitiy prediction and the third component, a MLP, for location prediction. The three parts are jointly trained end-to-end with a combined loss term, so the model is significantly smaller than pre-trained equivalents. %In contrast to approaches with pre-trained models, the authors highlight that their model is significantly smaller as it is only trained on data needed for the specific purpose.%
Mishra et al.~\cite{Mishra.2021} and Choi et al.~\cite{Choi.2022} advanced auto-encoder networks for AD by integrating them with a GMM and using a vision transformer encoder, respectively. Mishra et al.'s \textit{VT-ADL} model involves end-to-end training, feeding encoder output patch embeddings to a GMM for Gaussian calculations and a CNN decoder for reconstruction. They use mean squared error, log-likelihood and structural similarity (SSIM) losses as training objective and for the anomaly map generation. Choi et al.~\cite{Choi.2022} adopt a similar methodology but employ a variational auto-encoder and a pre-trained transformer, emphasizing a workflow from image collection to expert validation. Both approaches underscore transformers' superiority in capturing global dependencies over CNNs.
Fan et al.~\cite{Fan.2020} integrate GMMs with variational auto-encoders for pinpointing anomalies in surveillance video streams, employing a model that samples from multiple distributions to accommodate complex feature spaces.

Normalizing Flow models (NF) are located in two categories, generative models and representation based approaches and are an efficient alternative to GMMs for estimating complex distributions. They have seen increasing popularity for AD due to their fast inference and strong downstream performance.
They estimate the exact likelihood of features by following any arbitrary distribution and computing the likelihood by using the KL divergence between a prior and a base distribution~\cite{Gudovskiy.2021,Dinh.2016}.
Gudovskiy et al.~\cite{Gudovskiy.2021} introduce \textit{CFLOW-AD}, an AD model pairing a pre-trained \textit{ResNet} encoder with a conditional NF, training independently for each class with fixed position embeddings. 
% They report image-level AUROC and pixel-level PRO scores, using Euclidean distance from the mean to determine the optimal threshold for anomaly classification based on F1-score maximization.
Yu et al.~\cite{Yu.2021} critique fixed position embeddings for complex datasets, introducing \textit{FastFlow}, an AD model with a 2D NF that eliminates the need for positional encoding by preserving spatial structures. \textit{FastFlow} uses a \textit{ResNet} or a vision transformer encoder with selective embedding stages. The authors report AUROC scores on, among others, the MVTecAD and BTAD dataset, achieving speed gains over the previous models in~\cite{Gudovskiy.2021} and~\cite{Roth.6152021}. Lei et al.~\cite{Lei.352023} improve results with pyramid NFs and end-to-end trained $1\times1$ CNNs, while Kim et al.~\cite{Kim.10262022} enhance \textit{FastFlow} and \textit{CFLOW-AD} stability and performance with a novel training approach.

\iffalse
% reminder what we want to say
We provide a comprehensive overview of current state-of-the-art (SotA) approaches for the task of visual quality control
We benchmark a great variety of different anomaly detection methods with different visual backbones considering their performance in successfully detecting and localizing defective samples.
We recommend guidelines for choosing a suitable architecture for deploying a Quality Control System in practice considereing hardware constraints.

Considering the related work of our contributions, there are basically two different classes. The one are papers that propose an own approach for the AD and AL problem, as described in the last sections. The other are survey papers about the topic, for instance Tao et al.~\cite{Tao.2022}.
However, our work differentiates from both, since we not only survey describe the work in the field (like in a survey), but also implement and benchmark different approaches for AD and AL. Moreover, we propose the use of pre-trained vision transformers as visual backbones, which was not considered so far.
\fi

\section{Experimental Setup}
\label{sec:experiments}
\subsection{Backbone Architectures}
\label{sec:backbonearch}
Because of their performance advantages, our analysis primarily focuses on transformer-based architectures. Given the importance of hardware efficiency in practical applications, the lightweight and efficient hierarchical transformers are of particular interest in our study. For a comprehensive perspective, we also evaluate a classical \textit{ResNet-50} model as a baseline.  We chose this \textit{ResNet} variant since it is a common choice when making a trade-off between the quality of representations and computational cost \cite{Gudovskiy.2021,Yu.2021}.
The \textit{DeiT} architecture was chosen as an example of a monolithic vision transformer that closely follows the architecture of the original \textit{ViT} proposed in~\cite{Dosovitskiy.2020}, but achieves better performance by using knowledge distillation~\cite{Touvron.2020}. We used the largest variant \textit{DeiT-base} in our experiments since it has the highest performance and was also used in~\cite{Yu.2021}.
Moreover, we selected the \textit{EsViT} approach to investigate the performance of hierarchical transformer versions. We used the variant based on the \textit{Swin-T} architecture since it has a relatively small number of parameters which is comparable to \textit{ResNet-50}. A window size of 14 achieved the best results in~\cite{Li.6172021}, hence we adopted this choice for our experiments.
\textit{EfficientFormer} is another efficient transformer suitable for settings with limited computational resources.
To have a comparable parameter size to \textit{ResNet-50} and make a good trade-off between efficiency and performance, we picked the medium-sized model variant \textit{L3}.
A comparison of the number of parameters of the different image encoder models is shown in Table \ref{tab:parametersizeoverview}.
\begin{table}
    \centering
    \begin{tabular}{| l | ccc| c | c |}
        \hline
                                                &
        \multicolumn{3}{l|}{\textsc{Backbone} } &
        \textsc{GMM}                            &
        \textsc{NF}
        \\
                                                &
        \textsc{Pa}                             &
        \textsc{PE}                             &
        \textsc{FM}                             &
        \textsc{Pa}                       &
        \textsc{Pa}
        \\ [0.5ex]
        \hline
        \textsc{ResNet} \textbf{50}             &
        28M                                     &
        [512,1024,2048]                                    &
        [28,14,7]                                     &
        525M                                    &
        115M
        \\
        \hline
        \textsc{DeiT} \textbf{B}                &
        87M                                     &
        768                                     &
        14                                   &
        118M                                    &
        31M
        \\
        \hline
        \textsc{EsViT} \textbf{T}               &
        27M                                     &
        768                                     &
        7                                     &
        118M                                    &
        31M
        \\
        \hline
        \textsc{EffFormer}  \textbf{L3}         &
        31M                                     &
        512                                     &
        7                                     &
        53M                                     &
        14M
        \\
        \hline
    \end{tabular}
    \caption{Model configurations with parameter sizes (\textsc{Pa}) in million parameters, patch embedding size (\textsc{PE}) and size of the embedding feature maps (\textsc{FM}). The sizes of GMM and NF are dependent on the image backbone.}\label{tab:parametersizeoverview}
\end{table}
% architecture images (only include those if we alse write something about it here, might be duplicate with related work thogh)
\iffalse
\begin{figure}
    \centering
    \includegraphics[width=1\textwidth]{images/ArchitectureEsVitSwin.png}
    \caption{Architecture of the EsViT model~\cite{Li.6172021}.}\label{fig:archesvit}
\end{figure}
\begin{figure}
    \centering
    \includegraphics[width=1\textwidth]{images/ArchitectureEfficientFormer.png}
    \caption{Architecture of the EfficientFormer model~\cite{li2022efficientformer}.}\label{fig:archeffformer}
\end{figure}
\fi
\subsection{Anomaly Detection Architectures}
\label{sec:ad_architectures}
Based on the feature vector produced by the vision backbone a classifier needs to distinguish normal examples from defective anomalies. 
We focused our experiments on GMMs, which were one of the first approaches using a vision transformer backbone and NFs, which showed promising performance in AD tasks (as discussed in Section \ref{sec:ad_representation_based}).

\begin{figure}
    \centering
    \includegraphics[width=1\textwidth]{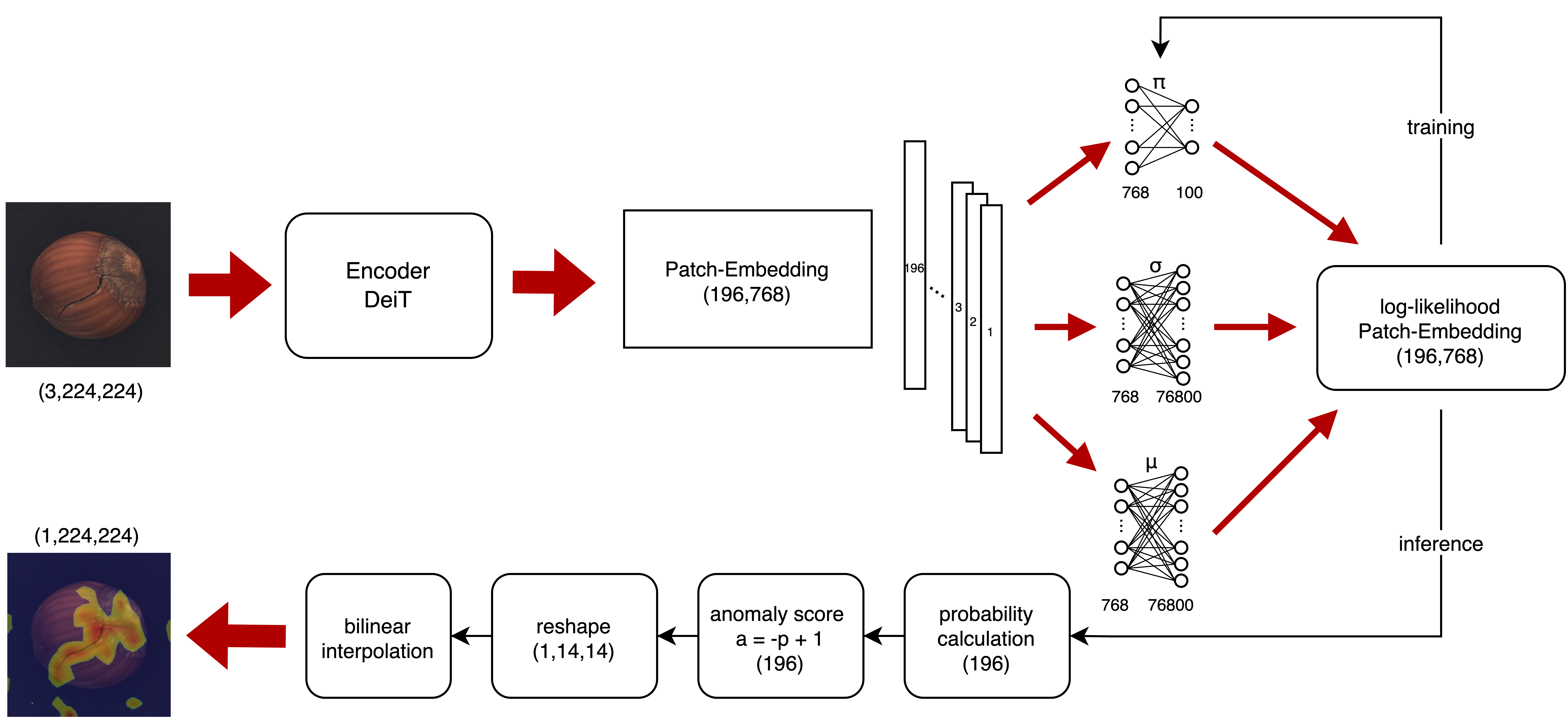}
    \caption{Gaussian Mixture Model in combination with \textit{DeiT} encoder.}\label{fig:archgmm}
\end{figure}

The performance of a GMM usually improves with a higher number of Gaussians but the use of a fully-connected MLP for every component can make them prohibitively expensive. Figure \ref{fig:archgmm} illustrates how we used image-processing backbones together with a GMM to perform AD in the case of \textit{DeiT}. The procedure for the other transformer backbones is analogous with differing numbers for the embedding size. With \textit{ResNet}, we followed the approach in \cite{Gudovskiy.2021} and trained two GMMs for the output of the last two blocks of \textit{ResNet-50} to capture global and local dependencies.
Based on the patch embeddings we trained three MLPs representing the means ($\mu$), standard deviations ($\sigma$) and weights ($\pi$) of the Gaussian Mixture components to capture the distribution of the normal examples. The size of the MLPs depends on the number of Gaussians. During training we minimized the negative log-likelihood, while for inference we calculated an anomaly score for every patch as follows: The log-likelihood is normalized to a value between zero and one using min-max normalization over the batch to obtain a pseudo-probability $p$. Our anomaly score is defined as $a = 1 - p$. To obtain a 2D anomaly map we reshaped the 196 patches to a $14\times14$ matrix and used bilinear interpolation to project the anomaly map to our original image size of $224\times224$. An image is deemed anomalous if its highest patch anomaly score surpasses a specific threshold, which is empirically determined using the validation set, as suggested by You et al.~\cite{You.952022}. 

\begin{figure}
    \centering
    \includegraphics[width=1\textwidth]{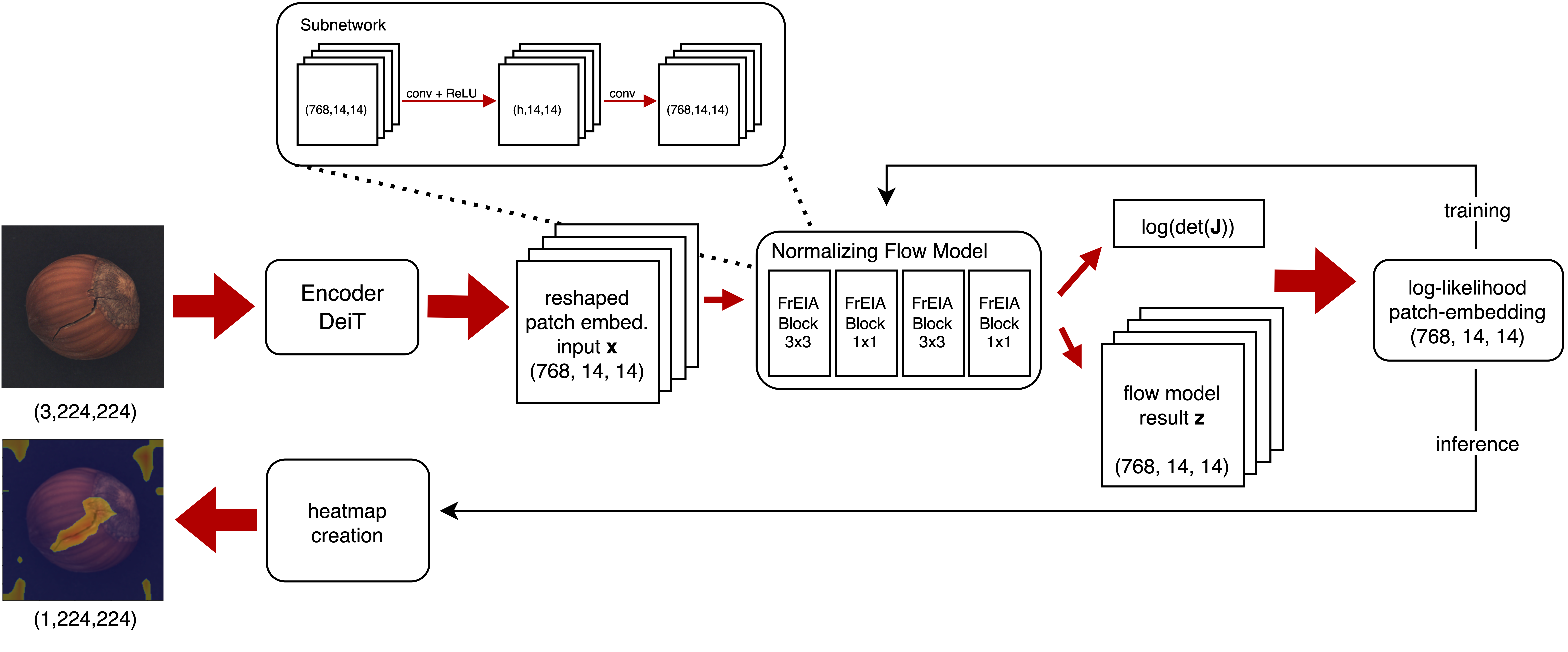}
    \caption{NF model with the \textit{DeiT} encoder. Each \textit{FrEIA} Block~\cite{freia} has a subnetwork with hidden dimension $h$, calculated with the hidden ratio from table \ref{tab:hyperparams}.}\label{fig:archnf}
\end{figure}

Normalizing Flows were chosen as models in view of limited compute. Our setup of a NF model combined with a \textit{DeiT} image-processing backbone is visualized in Figure \ref{fig:archnf}. The setup is analogous for other transformer backbones. For \textit{ResNet}, we used the output of the last three blocks and averaged the results, similar to the GMM approach.
The patch embeddings produced by the vision backbone are passed through the NF model consisting of 20 flow steps for the transformer-based methods and eight for \textit{ResNet-50}, due to hardware limitations.  Further, we chose a hidden ratio $h=0.16$. Each flow step consists of a subnetwork of alternating $3\times3$ and $1\times1$ convolutional layers. These hyperparameters follow the recommendations of~\cite{Yu.2021}. For the other hyperparameters see Table \ref{tab:hyperparams}. The flow steps were realized by \textit{AllInOneBlock}s of the \textit{FrEIA} framework~\cite{freia}. The NF model outputs the log determinant of the Jacobian and the transformed patches $z$. During training those two outputs are used for the loss objective. At inference, $\log|\operatorname{det} \boldsymbol{J}|$ and  $z$ are used to calculate the likelihood which is then used as anomaly map for localization, by upsampling the patches with bilinear interpolation. For more details on the NF architecture see~\cite{Yu.2021}.
% \begin{equation} \label{eq:cond_nf_loss}
%     \begin{aligned}
%         \mathcal{L}(\boldsymbol{\theta}) =D_{K L}\left[p_X(\boldsymbol{x}) \| \hat{p}_X(\boldsymbol{x}, \boldsymbol{c}, \boldsymbol{\theta})\right] \approx \frac{1}{N} \sum_{i=1}^N\left[\frac{\left\|\boldsymbol{z}_i\right\|_2^2}{2}-\log \left|\operatorname{det} \boldsymbol{J}_i\right|\right]+\mathrm{const}
%     \end{aligned}
% \end{equation}
% \iffalse
% \begin{itemize}
%     \item Architecture of Normalizing Flow Model as this is the best working approach
%     \item Architecture of GMM because this shows that pre-trained transformers are better than trained from scratch and that hierarchical transformers have potential
% \end{itemize}
% \fi
\subsection{Datasets}
We evaluate our methods on two datasets that are as close to real-life as possible.

The BTAD dataset consists of three different types of real-world industrial products and has a total of 2830 RGB images with a resolution of between $600\times600$ and $1600\times1600$~\cite{Mishra.2021}. Each product category is split into a train set, consisting of only normal images and a test set, which includes normal and anomalous images. In total there are 1799 training images, the rest are test images. The proportion of anomalous images in the test set varies from 9\% to 87\% depending on the class. A ground-truth mask for each anomalous image is provided that can be used for AL. There is no further distinction between body and surface defect anomaly classes in the labels.

The MVTecAD dataset consists of 15 different product categories or textures and has a total of 5354 images~\cite{bergmann.mvtecad}. Three classes are provided as grayscale images which is common in some real-life settings. The other classes are RGB images. The resolution varies between $700\times700$ and $1024\times1024$ and some classes have more examples than others.
Each product category is split into a training set with only normal instances and a test set that also contains anomalies. In total there are 3629 training and 1725 test examples. The proportion of abnormal images is, in most cases, about two to three times higher than the normal ones in the test set. The abnormal data is classified into different types of anomalies, for example in crack, hole, cut and print on the \textit{hazelnut} class. A ground-truth mask on pixel-level is given for each abnormal image.

\subsection{Implementation Details}
Since both datasets do not provide a separate validation set for hyperparameter tuning and model selection we split the training set into 80\% train and 20\% validation data in both cases. We applied Min-Max scaling to every image. The minimum and maximum values were computed separately for every channel on the training set. We scaled the input images to an image size of $224\times224$ pixels which is the default training size of most transformer models \cite{Dosovitskiy.2020}. 

Hyperparameters were optimized once for each model using the \textit{hazelnut} class of the MVTecAD dataset and the best configuration according to validation loss is adopted for all other experiments. We chose this class because it has the largest amount of training samples in the dataset. A summary of the considered hyperparameters and the best values found for them are shown in Table \ref{tab:hyperparams}. Note that the batch size of the GMM is relatively small due to hardware limitations.
We trained each model separately on every object class for a maximum of 500 epochs using early stopping based on the validation loss with a patience of 30 epochs. Only the parameters of the AD models were updated, the image-backbones were pre-trained on ImageNet1k~\cite{deng2009imagenet} and kept frozen during training. For evaluation the model checkpoint with the best validation loss was selected.
All models were trained on a single NVIDIA GeForce RTX 2080 Ti TURBO GPU with 11GB VRAM.

\begin{table}
    \centering
    \begin{tabular}{| p{0.1\textwidth} | p{0.15\textwidth} | p{0.1\textwidth} | p{0.1\textwidth} | p{0.1\textwidth} | p{0.15\textwidth} | p{0.1\textwidth} | p{0.1\textwidth} |}
        \hline
        \textsc{ad type}             &
        \textsc{encoder}             &
        \textsc{batch size}          &
        \textsc{learn rate}       &
        \textsc{weight decay}        &
        \textsc{number of gaussians} &
        \textsc{flow steps}          &
        \textsc{hidden ratio}
        \\ [0.5ex]
        \hline
        \textsc{gmm}                     &
        ResNet50                     &
        4                            &
        1e-4                         &
        1e-4                         &
        50                           &
        -                            &
        -
        \\
                                     &
        Other                        &
        8                            &
        1e-4                         &
        1e-4                         &
        100                          &
        -                            &
        -
        \\
        \hline
        \textsc{nf}                          &
        ResNet50                     &
        16                           &
        1e-4                         &
        1e-5                         &
        -                            &
        8                            &
        0.16
        \\
                                     &
        EsViT                        &
        32                           &
        1e-4                         &
        1e-5                         &
        -                            &
        20                           &
        0.16
        \\
                                     &
        Other                        &
        32                           &
        1e-3                         &
        1e-5                         &
        -                            &
        20                           &
        0.16
        \\
        \hline
    \end{tabular}
    \caption{Hyperparameter configuration of the different models architectures.}\label{tab:hyperparams}
\end{table}
% comments on specific combinations of backbone + AD method
% \iffalse
% \paragraph{ResNet + NF}
% - already evaluated from the Authors of the NF Paper, just reproduced
% \paragraph{DeiT + NF}
% - already evaluated from the Authors of the NF Paper, just reproduced
% \paragraph{EsViT + NF}
% - is using swin Transformer, often mentioned in terms of segmentation tasks (where?)
% \paragraph{EfficientFormer + NF}
% - V2 is available, should we retry some experiments?
% \paragraph{Deit + GMM}
% - outperformed vtadl in some classes
% \paragraph{EsViT + GMM}
% - outperformed FastFlow in combination with EsVit but not the same classes were used
% \paragraph{EfficientFormer + GMM}
% \fi
\subsection{Metrics}
For the evaluation of our approach, we use the AUROC since it is a common metric for visual quality control~\cite{Yu.2021,Bae.11222022,Roth.6152021,Hyun.5262023}. For evaluating AD we consider the AUROC on image and for AL on pixel level.

As a second metric we use the Per-Region-Overlap (PRO), which measures the overlap between ground truth and predicted anomalies on a pixel level. To calculate the PRO-score one needs to set a threshold at which level to classify a pixel as anomalous. This is usually done based on the ROC considering a trade-off between acceptable levels of False Positive Rates (FPR) and TPRs. This can depend on the use case, considering whether it is costly to allow a high FPR (e.g. if a false positive results in an entire production batch being discarded). All pixels with anomaly values below the threshold are set to zero, while the rest keep their original anomaly score.
We calculate the area under the PRO-curve up to a maximum false positive rate of 30\% following the procedures of~\cite{Bergmann.2020} and~\cite{Mishra.2021} for the same datasets.
We use the PRO-score to evaluate AL.

Furthermore, we report the precision recall area under curve (PRAUC) as supplement to the AUROC for a more detailed view of the overall performance and to compare results with other works that also use this score. PRAUC is only used to measure image-level AD performance.

\subsection{Experiments}
\label{subsec:experiments}
Our first goal was to assess if similar outcomes can be achieved as the \textit{VT-ADL} architecture by Mishra et al.~\cite{Mishra.2021}. While they trained a vision transformer with a GMM and a CNN from scratch on the BTAD and MVTecAD datasets, we used a pre-trained \textit{DeiT} model as a frozen image backbone and trained only a GMM with the likelihood loss. Chosen for its enhanced performance over \textit{ViT}, our \textit{DeiT} configuration includes twelve layers and twelve heads, versus the six layers and eight heads used by \textit{VT-ADL}.
We used an image resolution of $224\times224$ instead of $500\times500$ to match the requirements of our pre-trained image encoders and employed a smaller GMM with only 100 instead of 150 Gaussians according to our empirical hyperparameter search.
%We used a streamlined GMM of 100 instead of 150 Gaussians, reducing the model size from 177 million to 118 million parameters and employing a 224x224 image resolution instead of 500x500.
Additionally, we attempted to reproduce Yu et al.'s~\cite{Yu.2021} promising \textit{FastFlow} results for MVTecAD, despite the absence of their source code. We followed their experimental details. While the authors use the 7th \textit{DeiT} encoder block's embedding we additionally evaluate a model version which uses \textit{DeiT}'s last layer. For compatability with our pre-trained backbones we scale the images to a size of $224\times224$ instead of $384\times384$ pixels.

A second goal was to study the behavior of GMMs and NFs on a more fine-grained level to find out if there are differences in the performance depending on the object class. For this experiment we used our setup with a pre-trained \textit{DeiT} image encoder and trained and evaluated the 15 classes of the MVTecAD dataset separately. For the NF model we investigated two different versions, NF i11 which uses the features from the last layer of the image encoder and NF i7 which uses the feature maps of the 7th encoder layer. For the sake of completeness we also compared the performance with the reported results from~\cite{Mishra.2021} and from~\cite{Yu.2021}. 
%The authors do not report their AD performance and Yu et al. do only report the mean performance but not the class level performance.
% Second, we study the behavior of GMM and NF on a more fine-grained level to find out if there is differences in performance depending on the object class. For this experiment we use our setup with a pre-trained Deit image encoder and train and evaluate the 15 classes of the MVTechAD dataset separately. For the NF model we investigate two different versions NF i11 which uses the features from the 11th layer of the image encoder and NF i7 which uses the feature maps of the 7th encoder layer. The latter was proposed in \cite{Yu.2021}. For completenes sake we also compare the class-level AL performance with the results from \cite{Mishra.2021}. The authors do not report their AD performance and Yu et al. do only report the mean performance but not the class level performance.
\begin{table}
    \centering
    \begin{tabular}{|l|cc|cccc|cccc|} \hline
                            & \textsc{vt-adl}         &                    & \textsc{gmm}            &                    & \textsc{image} & \textsc{pixel} & \textsc{nf i11}            &                    & \textsc{image} & \textsc{pixel} \\
        \textsc{data class} & \textsc{prauc}          & \textsc{pro} & \textsc{prauc}          & \textsc{pro} & \textsc{auroc}          & \textsc{auroc} & \textsc{prauc}          & \textsc{pro} & \textsc{auroc}          & \textsc{auroc}          \\ \hline
        1                   & \textbf{99.00} & \textbf{92.00}     & 97.25          & 78.83              & 93.25          & 81.68          & 99.96 & 80.59 & \textbf{99.90} & \textbf{84.55} \\
        2                   & 94.00          & 89.00              & \textbf{95.50} & \textbf{92.22}     & 75.29          & 92.20          & 96.60 & 87.58 & \textbf{79.36} & 87.85 \\
        3                   & \textbf{77.00} & 86.00              & 19.10          & \textbf{90.49}     & 67.18          & 91.96          & 96.85 & 95.68 & \textbf{99.65} & \textbf{95.72} \\ \hline
        mean                & \textbf{90.00} & \textbf{89.00}     & 70.62          & 87.18              & 78.57          & 88.61          & 97.80 & 87.95 & \textbf{92.97} & \textbf{89.37} \\ \hline
    \end{tabular}
    \caption{Results on BTAD of \textit{VT-ADL} [24] and GMM and NF i11 with \textit{DeiT}.}\label{tab:comparisonVTADLonBtad}
\end{table}

Third, we examined the performance effects of replacing traditional monolithic transformers with hierarchical versions using the \textit{EsVit} and the \textit{EfficientFormer} models described in Section \ref{sec:backbonearch}. We also consider the CNN-based \textit{ResNet-50} backbone as a baseline.  As proposed in \cite{Gudovskiy.2021} and discussed in Section~\ref{sec:ad_architectures}, we used the output of several blocks of the \textit{ResNet} backbone and averaged the results. In case of the GMM, we trained only two models with 50 Gaussians for the last two blocks, due to hardware limitations.
We trained and evaluated the methods on five selected classes of the MVTecAD dataset. The classes were chosen separately for both the GMM and the NF i11 approaches based on the performance of our previous experiment with the \textit{DeiT} encoder. As a representative sample for each model, we included classes with high, medium and low performance according to our experiments. For GMM these are the classes \textit{cable}, \textit{carpet}, \textit{grid}, \textit{hazelnut} and \textit{tile}. For the NF i11 we use the classes \textit{bottle}, \textit{carpet}, \textit{hazelnut}, \textit{leather} and \textit{screw}.
\section{Results and Discussion}
\label{sec:results}
\subsection{Comparison with the VT-ADL and FastFlow models}
\begin{figure}[ht]
    \centering
    \begin{subfigure}[t]{0.15\textwidth}
        \includegraphics[height=0.85in]{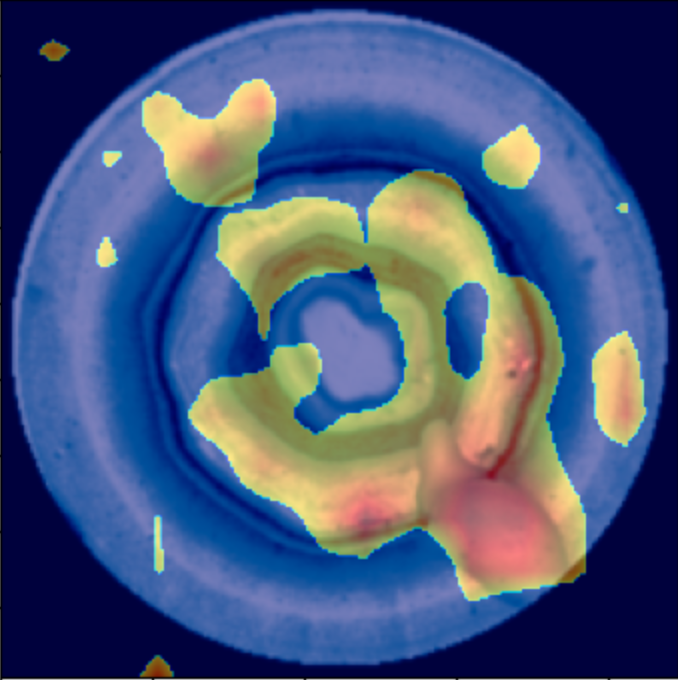}
    \end{subfigure}
    \hspace{0.5em}
    \begin{subfigure}[t]{0.15\textwidth}
        \includegraphics[height=0.85in]{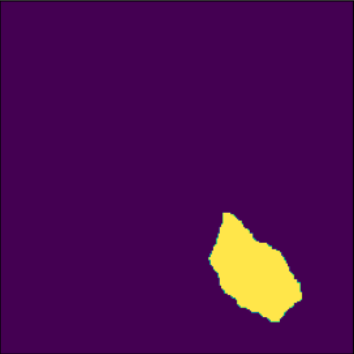}
    \end{subfigure}
    \hspace{0.5em}
    \begin{subfigure}[t]{0.15\textwidth}
        \includegraphics[height=0.85in]{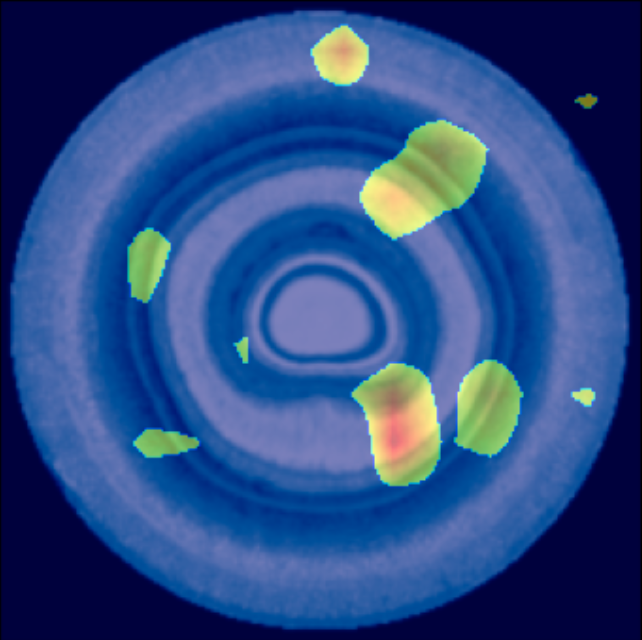}
    \end{subfigure}
    \hspace{0.5em}
    \begin{subfigure}[t]{0.15\textwidth}
        \includegraphics[height=0.85in]{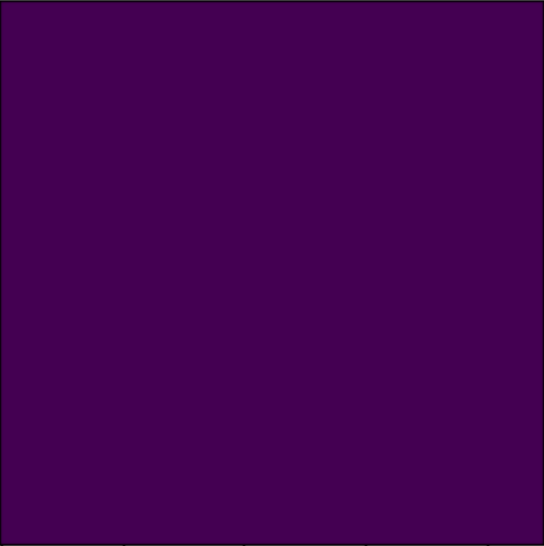}
    \end{subfigure}
    \caption{The third class of the BTAD dataset, processed by the GMM with \textit{DeiT}. Anomaly map and ground truth for a defective (left) and normal (right) sample.}\label{fig:btad03gmm}
\end{figure}

Table \ref{tab:comparisonVTADLonBtad} shows that our GMM model performs better than \textit{VT-ADL} in two of three classes of the BTAD dataset in the localization task and one class in the detection class. The bad detection performance on class three may result from the model highlighting anomalies correctly but also producing spots with high anomaly scores in normal images (see Figure \ref{fig:btad03gmm}). This results in a high false positive rate when using the maximum of the patch anomaly scores to classify the image. Considering also the composition of the test dataset for this class can explain the gap between detection and localization performance, since it has about ten times more normal than anomalous samples. The overall localization performance of our model on the MVTecAD dataset is higher than the one of \textit{VT-ADL}. These results show, that using a pre-trained transformer encoder in scenarios with relatively small datasets can be beneficial compared to training a transformer end-to-end. 

Table \ref{tab:MVTECallMethods} shows, that our implementation of the NF model could not reach the values reported for \textit{FastFlow} in \cite{Yu.2021}. However, the use of only 80\% of the training data and a lower image size may have negatively affected the performance.
\subsection{Comparison of GMMs and NF models}
\begin{table}
    \centering
\begin{tabular}{|l|cc|cc|cc|c|cc|} \hline
    architecture & \multicolumn{2}{l|}{\textsc{nf i11}} & \multicolumn{2}{l|}{\textsc{nf i7}} & \multicolumn{2}{l|}{\textsc{gmm}} & \textsc{vt-adl} & \multicolumn{2}{l|}{\textsc{FastFlow}} \\
    dataclass & \textsc{ad} & \textsc{al} & \textsc{ad} & \textsc{al} & \textsc{ad} & \textsc{al} & \textsc{al} & \textsc{ad} & \textsc{al(auroc)} \\ \hline
        bottle & \textbf{99.84} & 85.95 & 98.57 & \textbf{93.64} & 98.49 & 45.24 & 94.90 & 100.00 & 97.70 \\
        cable & \textbf{94.42} & \textbf{94.68} & 81.73 & 84.87 & 61.94 & 78.03 & 77.60 & 100.00 & 98.40 \\
        capsule & \textbf{92.94} & 93.42 & 92.82 & \textbf{95.04} & 82.69 & 93.69 & 67.20 & 100.00 & 99.10 \\
        carpet & 96.07 & 95.92 & 87.40 & 87.40 & \textbf{97.11} & \textbf{97.06} & 77.30 & 100.00 & 99.40 \\
        grid & \textbf{98.08} & 93.79 & 96.66 & \textbf{96.11} & 76.44 & 71.21 & 87.10 & 99.70 & 98.30 \\
        hazelnut & \textbf{98.46} & 95.81 & 87.18 & 93.06 & 65.43 & \textbf{96.86} & 89.70 & 100.00 & 99.10 \\
        leather & \textbf{100.00} & 88.10 & 99.97 & \textbf{98.83} & 98.37 & 98.62 & 72.80 & 100.00 & 99.50 \\
        metal nut & \textbf{99.71} & \textbf{92.98} & 87.93 & 88.69 & 60.95 & 84.54 & 72.60 & 100.00 & 98.50 \\
        pill & \textbf{92.17} & \textbf{94.25} & 81.29 & 90.83 & 69.01 & 91.35 & 70.50 & 99.40 & 99.20 \\
        screw & \textbf{86.53} & \textbf{95.96} & 77.09 & 84.00 & 52.43 & 46.37 & 92.80 & 97.80 & 99.40 \\
        tile & \textbf{99.96} & 91.43 & 96.03 & \textbf{93.47} & 92.50 & 90.38 & 79.60 & 100.00 & 96.30 \\
        toothbrush & 90.83 & 93.50 & 88.61 & 95.24 & \textbf{95.56} & \textbf{95.84} & 90.10 & 94.40 & 98.90 \\
        transistor & \textbf{95.71} & \textbf{96.52} & 88.21 & 93.29 & 76.83 & 92.31 & 79.60 & 99.80 & 97.30 \\
        wood & 92.46 & 87.73 & \textbf{98.25} & \textbf{93.86} & 93.33 & 90.66 & 78.10 & 100.00 & 97.00 \\
        zipper & \textbf{96.68} & 90.91 & 92.67 & \textbf{97.31} & 79.28 & 95.81 & 80.80 & 99.50 & 98.70 \\ \hline \hline
        mean & \textbf{95.59} & \textbf{92.73} & 90.29 & 92.38 & 80.02 & 84.53 & 80.71 & 99.37 & 98.45 \\
        std & \textbf{4.00} & \textbf{3.28} & 6.92 & 4.38 & 15.49 & 17.41 & 8.18 & 1.44 & 0.94 \\ \hline
    \end{tabular}
    \caption{Image AUROC (AD) and PRO (AL) of the NF model with feature maps from last and seventh \textit{DeiT} block, the GMM, \textit{VT-ADL} and \textit{FastFlow}. Standard deviation (std) is calculated across classes. \textit{FastFlow} reported pixel AUROC.}
    \label{tab:MVTECallMethods}
\end{table}
Table \ref{tab:comparisonVTADLonBtad} and \ref{tab:MVTECallMethods} show that the overall performance of the NF model is better than the GMM. However, the GMM performs better in localization tasks on most of the surface classes. The high standard deviation between the classes on MVTecAD for the GMM is mainly caused by the classes \textit{bottle} and \textit{screw}, which could not be learned correctly.
The NF-model has a relatively small standard deviation between the classes and thus is more robust when applied to new classes. Together with the smaller size this makes it more suitable for a use in production scenarios. The weakness of NF i11 in localization performance on some classes can be eliminated when using NF i7 instead (see Table \ref{tab:MVTECallMethods}). However, the overall detection performance of NF i7 is about five percent points less. The anomaly maps show that when using the last block, the highlighted areas are more coarse grained but the model is also more certain. Using the seventh block highlights anomalies more precisely but is also more uncertain and produces small anomaly spots on anomaly-free images. 
Regarding the model size (see Table \ref{tab:parametersizeoverview}), the NF model has a clear advantage over the GMM.
\subsection{Performance of the Backbones}
\begin{figure}[ht]
    \centering
    \begin{subfigure}[t]{0.15\textwidth}
        \includegraphics[height=0.85in]{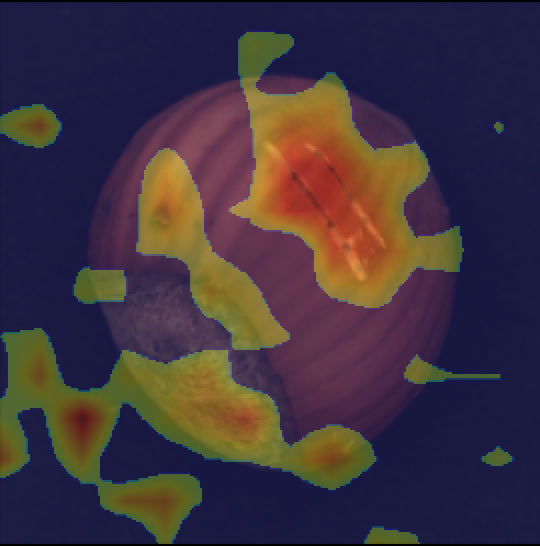}
    \end{subfigure}
    \hspace{0.5em}
    \begin{subfigure}[t]{0.15\textwidth}
        \includegraphics[height=0.85in]{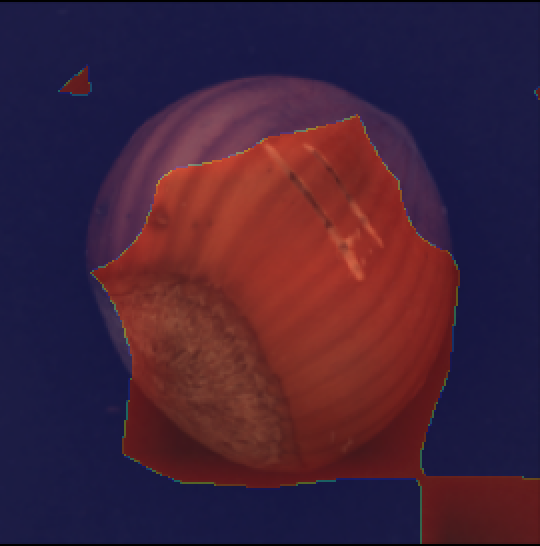}
    \end{subfigure}
    \hspace{0.5em}
    \begin{subfigure}[t]{0.15\textwidth}
        \includegraphics[height=0.85in]{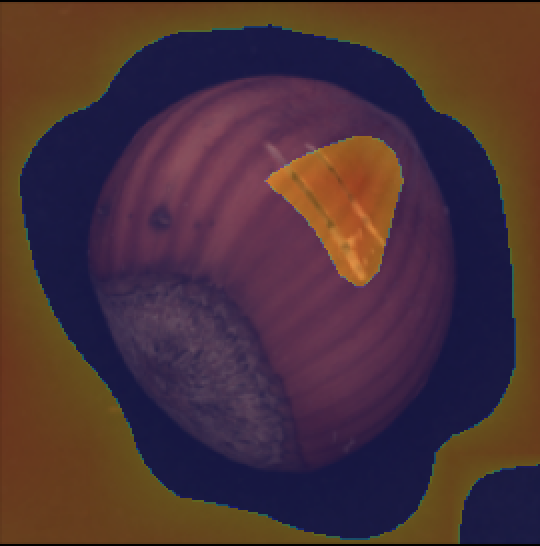}
    \end{subfigure}
    \hspace{0.5em}
    \begin{subfigure}[t]{0.15\textwidth}
        \includegraphics[height=0.85in]{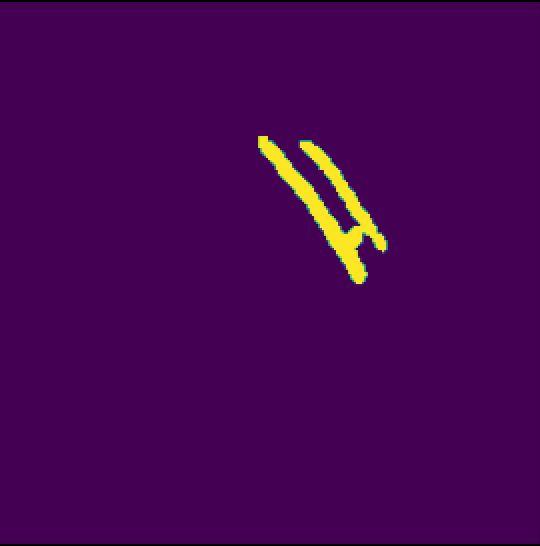}
    \end{subfigure}
    
    \begin{subfigure}[t]{0.15\textwidth}
        \includegraphics[height=0.85in]{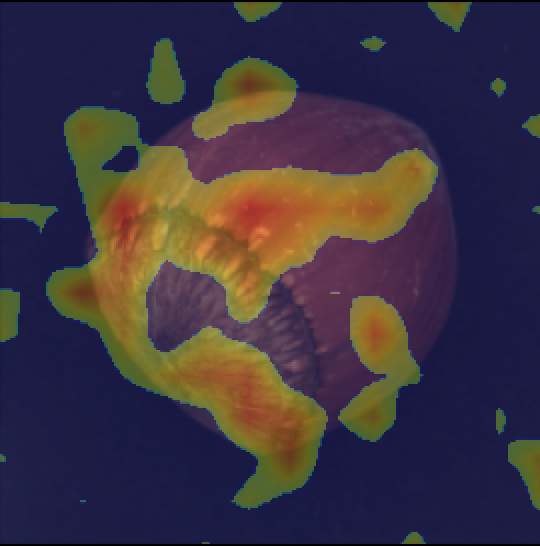}
    \end{subfigure}
    \hspace{0.5em}
    \begin{subfigure}[t]{0.15\textwidth}
        \includegraphics[height=0.85in]{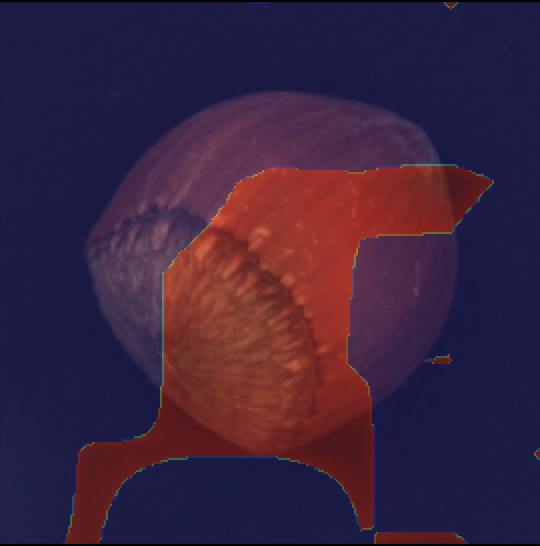}
    \end{subfigure}
    \hspace{0.5em}
    \begin{subfigure}[t]{0.15\textwidth}
        \includegraphics[height=0.85in]{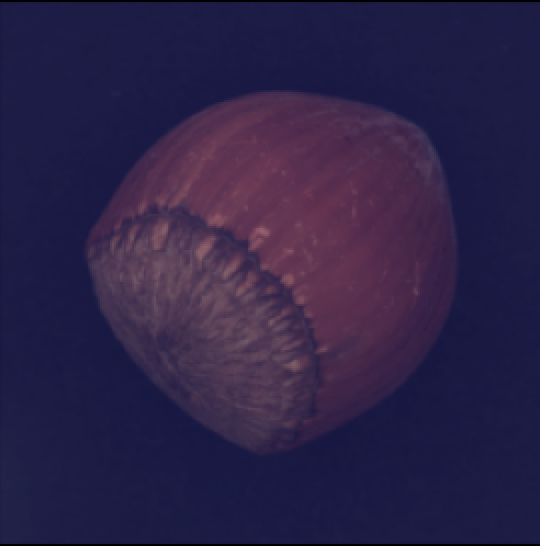}
    \end{subfigure}
    \hspace{0.5em}
    \begin{subfigure}[t]{0.15\textwidth}
        \includegraphics[height=0.85in]{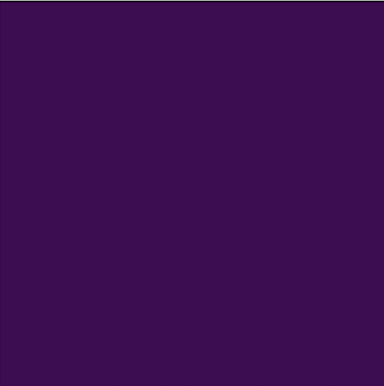}
    \end{subfigure}
    \caption{Anomaly maps for the \textit{hazelnut} class of a GMM with the backbones \textit{DeiT}, \textit{EfficientFormer}, \textit{EsViT} and the corresponding ground truth.}\label{fig:hazelnutheatmap}
\end{figure}

\begin{table}
    \centering
    \begin{tabular}{|l|lccc|lccc|} \hline
    & \multicolumn{4}{l|}{\textsc{auroc ad}} & \multicolumn{4}{l|}{\textsc{pro score}} \\
        dataclass & DeiT & EffFormer & EsViT & ResNet & DeiT & EffFormer & EsViT & ResNet \\ \hline
        cable & 62.00 & 52.00 & \textbf{89.00} & 79.00 & 78.00 & \textbf{81.00} & 72.00 & 61.00 \\
        carpet & \textbf{97.00} & 78.00 & 93.00 & 74.00 & \textbf{97.00} & 86.00 & 70.00 & 58.00 \\
        grid & 66.00 & \textbf{78.00} & 75.00 & 68.00 & \textbf{72.00} & 71.00 & 61.00 & 54.00 \\
        hazelnut & 66.00 & 54.00 & \textbf{97.00} & 73.00 & \textbf{97.00} & 88.00 & 68.00 & 51.00 \\
        tile & 96.00 & 74.00 & \textbf{99.00} & 65.00 & \textbf{90.00} & 74.00 & 49.00 & 54.00 \\ \hline
        mean & 77.40 & 67.20 & \textbf{90.60} & 71.80 & \textbf{86.80} & 80.00 & 64.00 & 55.60 \\ \hline
    \end{tabular}
\caption{AD and AL score of different backbones for the GMM on MVTecAD.}  \label{tab:diffbackbonesgmm}
\end{table}
Table \ref{tab:diffbackbonesgmm} shows that for the GMM the overall localization performance is the best with the \textit{DeiT} backbone. It outperforms the hierarchical backbones on all classes except for the \textit{cable} class. In contrast, the \textit{EsViT} model performs best in three of five classes in the detection task and has the overall best detection performance. A possible reason for the gap between localization and detection performance can be seen on the generated anomaly maps in Figure \ref{fig:hazelnutheatmap}. The \textit{EsViT} model does locate the anomaly correctly but also highlights large areas in the background. On the normal image there is no area highlighted at all. A possible reason for the bad performance of the \textit{ResNet} backbone is the usage of only 50 Gaussians and two output layers as discussed in Section \ref{subsec:experiments}.

\begin{figure}[ht]
    \centering
    \begin{subfigure}[t]{0.15\textwidth}
        \includegraphics[height=0.85in]{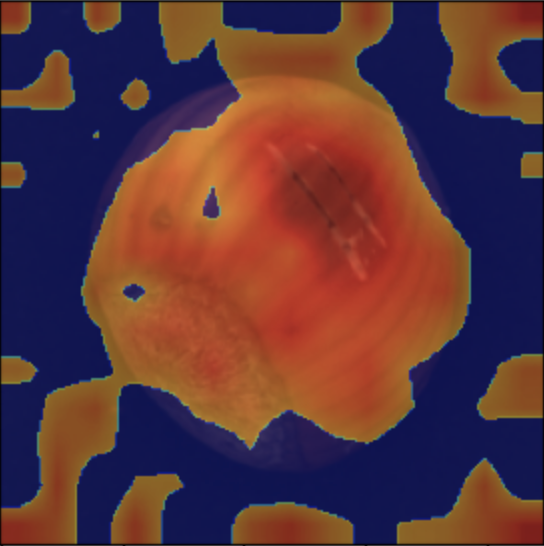}
    \end{subfigure}
    \hspace{0.5em}
    \begin{subfigure}[t]{0.15\textwidth}
        \includegraphics[height=0.85in]{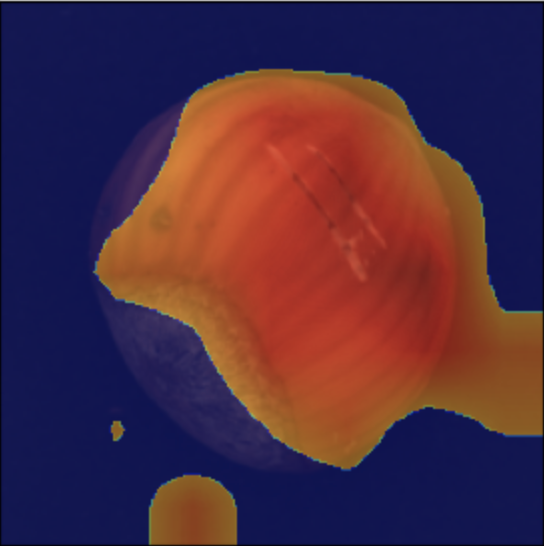}
    \end{subfigure}
    \hspace{0.5em}
    \begin{subfigure}[t]{0.15\textwidth}
        \includegraphics[height=0.85in]{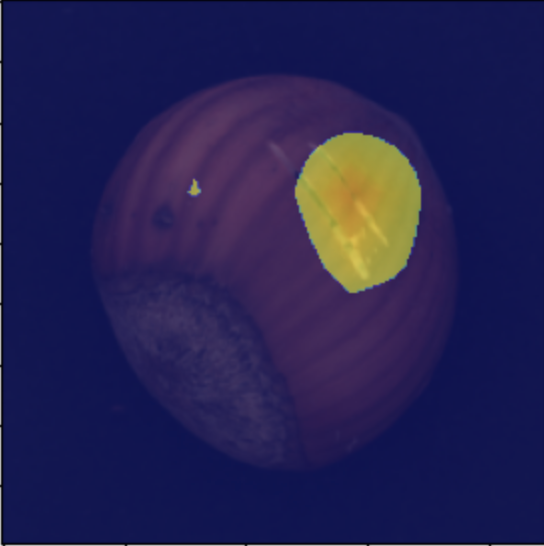}
    \end{subfigure}
    \hspace{0.5em}
    \begin{subfigure}[t]{0.15\textwidth}
        \includegraphics[height=0.85in]{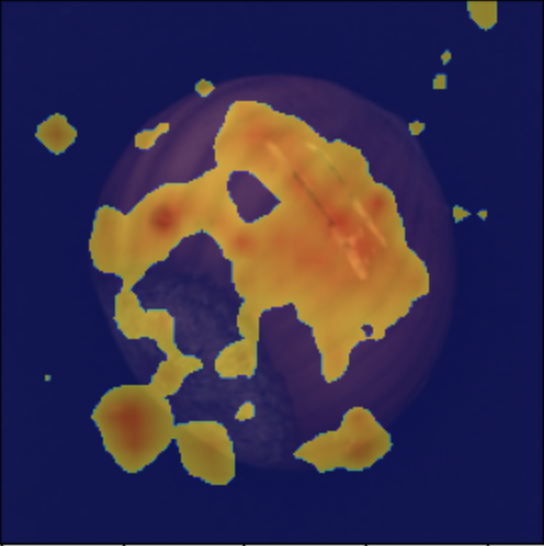}
    \end{subfigure}
    \hspace{0.5em}
    \begin{subfigure}[t]{0.15\textwidth}
        \includegraphics[height=0.85in]{images/hazelnut_gt_broken.png}
    \end{subfigure}
    \begin{subfigure}[t]{0.15\textwidth}
        \includegraphics[height=0.85in]{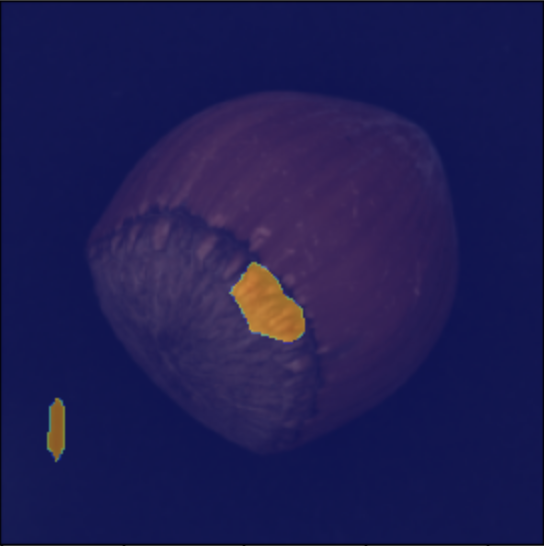}
    \end{subfigure}
    \hspace{0.5em}
    \begin{subfigure}[t]{0.15\textwidth}
        \includegraphics[height=0.85in]{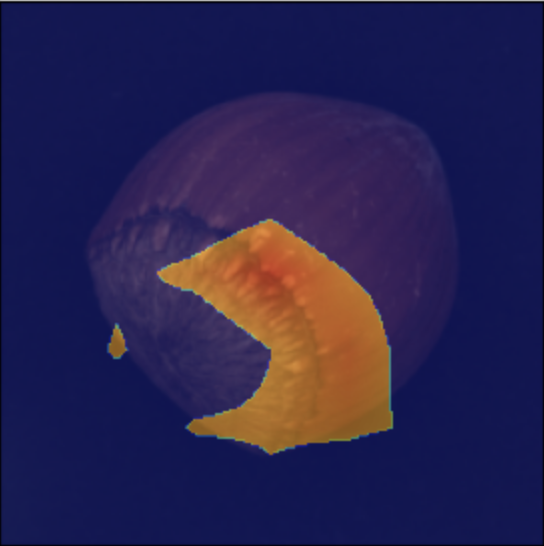}
    \end{subfigure}
    \hspace{0.5em}
    \begin{subfigure}[t]{0.15\textwidth}
        \includegraphics[height=0.85in]{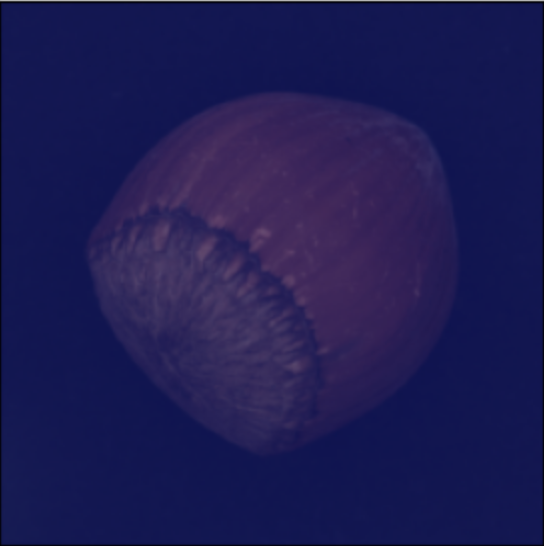}
    \end{subfigure}
    \hspace{0.5em}
    \begin{subfigure}[t]{0.15\textwidth}
        \includegraphics[height=0.85in]{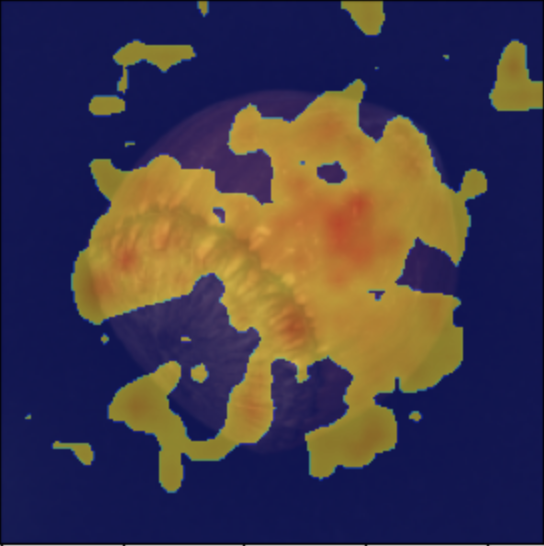}
    \end{subfigure}
    \hspace{0.5em}
    \begin{subfigure}[t]{0.15\textwidth}
        \includegraphics[height=0.85in]{images/gt_good.png}
    \end{subfigure}
    \caption{Anomaly maps for the \textit{hazelnut} class of the NF model with the backbones \textit{DeiT}, \textit{EfficientFormer}, \textit{EsViT}, \textit{ResNet} and the corresponding ground truth.}\label{fig:hazelnutheatmapnf}
\end{figure}

The results in Table \ref{tab:diffbackbonesnf} show, that \textit{DeiT} is the backbone that results in the best detection performance for the NF model. Nevertheless, \textit{EsVit} follows with the second best detection performance. 
Interestingly, while in general the NF achieved the best results, for \textit{EsViT} the GMM resulted in a better performance.
%and in combination with the GMMs, it has an overall better performance than the combination of \textit{EsVit} with the NF model. 
The \textit{ResNet} backbone outperforms \textit{DeiT} in two localization tasks but performs worse in detection tasks. Figure \ref{fig:hazelnutheatmapnf} shows the anomaly maps generated with the different backbones.

Hierarchical transformers create smaller models than CNNs, producing compact, information-rich patch embeddings, as highlighted in Table \ref{tab:parametersizeoverview}, where the patch embedding size significantly affects the GMM and NF sizes. Unlike \textit{DeiT}, these models are also smaller or equal in size to \textit{ResNet}.
For high-resolution images, the adequacy of transformer patch embedding for identifying small anomalies needs evaluation. Smaller embeddings can result in coarser feature maps due to upsampling from the patch embedding level.
\begin{table}
    \centering
    \begin{tabular}{|l|lccc|lccc|} \hline
        & \multicolumn{4}{l|}{\textsc{auroc ad}} & \multicolumn{4}{l|}{\textsc{pro score}} \\
        dataclass & DeiT & EffFormer & EsVit & ResNet & DeiT & EffFormer & EsVit & ResNet \\ \hline
        bottle & \textbf{100.00} & 98.00 & 93.00 & 97.00 & 86.00 & 79.00 & 66.00 & \textbf{97.00} \\
        carpet & \textbf{96.00} & 78.00 & 87.00 & 89.00 & \textbf{96.00} & 78.00 & 82.00 & 88.00 \\
        hazelnut & \textbf{98.00} & 89.00 & 93.00 & 91.00 & \textbf{96.00} & 92.00 & 73.00 & 90.00 \\
        leather & \textbf{100.00} & 77.00 & 96.00 & \textbf{100.00} & 88.00 & 85.00 & 63.00 & \textbf{97.00} \\
        screw & \textbf{87.00} & 43.00 & 65.00 & 50.00 & \textbf{96.00} & 60.00 & 57.00 & 84.00 \\ \hline
        mean & \textbf{96.20} & 77.00 & 86.80 & 85.40 & \textbf{92.40} & 78.80 & 68.20 & 91.20 \\ \hline
    \end{tabular}
    \caption{AD and AL score of different backbones for the NF model on MVTecAD.}
    \label{tab:diffbackbonesnf}
\end{table}

\subsection{Considerations and Limitations for Practical Application}
Anomaly maps can be used in various manufacturing scenarios such as explanation of a model's choice, making decisions based on the anomaly size or to double check a model's decision~\cite{Choi.2022}. However, it is important to notice that the experiments in this work are conducted on benchmark datasets that have high quality, which is hard to achieve in a real-world scenario. In manufacturing, metrics should be selected considering the composition of training data and the severity of false positives and false negatives. These aspects should also be considered when deciding on the thresholding strategy for the PRO-score and for AD. Our experiments on the detection performance show, that AUROC score should be preferred when false-negatives are expensive while PRAUC is more suitable when false-positives lead to problems. To evaluate the quality of a model when facing highly imbalanced datasets, both scores should be considered. For localization tasks, the PRO-score should be preferred. The model selection is highly influenced by the available data, hardware and real-time requirements. If enough hardware is available, the superior performance of \textit{DeiT} or a comparable monolithic transformer should be chosen. In case of strong hardware limitations and/or real-time requirements, smaller and faster hierarchical transformer are the models of choice. Yu et al. report an inference time of eight milliseconds with \textit{DeiT} and their \textit{FastFlow} model~\cite{Yu.2021}, what can be regarded as a baseline for hierarchical transformer models. When processing images with a much higher resolution, adjusting the patch-size might be required and thus a training of the backbone becomes necessary. 
%MLOps approaches should be consinderd to manage re-training  depending on the frequency of data change. In this case the training time of the model gets more important.

\section{Conclusion}
In our work, we provided a comprehensive overview of SoTA vision transformer models and evaluated different paradigms for visual anomaly detection in industrial visual quality control. We implemented two anomaly detection methods with four different image encoding backbones, all of them pre-trained on ImageNet1k. We trained our anomaly detection models on the datasets MVTecAD and BTAD and compared our results with two existing approaches from the literature. Finally, we discussed important aspects to consider when applying these approaches to production. We showed that using transformer models can improve the performance of anomaly detection models and reduce the overall size compared to \textit{ResNet}. Moreover, we showed that using pre-trained transformer models can have an advantage over training from scratch. Hierarchical transformer models are worth to evaluate for further use in production scenarios with limited computational capacity. Future research could perform experiments with using the output of more than one layer of the encoder model as done with \textit{ResNet} and proposed in~\cite{Xie.5312021} for their segmentation transformer.

% uncommented aspects I want to highlight. Keep in mind to proove that they are represented in the paper
% - When using deep learning with normalizing flow models with monolithic vision transformer models, the overall model size is reduced by about one quarter, even though the encoder model is three times as big as resnet, just because the decoder model size is reduced to nearly a quarter of the original size. This is due to the smaller emebedding created by the encoder and the fact that when using resnet it is neccessary to use the embedding of three encoder stages while with the transformer the embedding of one stage is enough

% - using hierarchical transformers can decrease the decoder size up to eight times, from 115M Parameter to about 14M Parameter. Unfortunately the results in combination with normalizing flow models were not competitive at the moment. However, transformer technology can be a chance for creating powerful models that create a small embedding with much information

% - using GMMs deliver worse results and bigger model. Thus they are not suitable for the current use case. However, using GMMs with EsVit increases the performance of the model even to a better one than when using NFs with EsVit. This theory has to be validated since the set of classes differs in the experiments with GMMs and NFs.
\begin{credits}
    \subsubsection{\ackname} This preprint has not undergone peer review or any post-submission improvements or corrections. The Version of Record of this contribution is published in „Machine Learning and Knowledge Discovery in Databases. Applied Data Science Track“ and is available online at \url{https://doi.org/10.1007/978-3-031-70381-2}.
    \newline Christoph Hönes has received funding from SAP SE. Christoph Hönes and Miriam Alber were employed by esentri AG who also provided computational resources.
\end{credits}
\bibliographystyle{splncs04}
\bibliography{references}




% LNCS Homepage, \url{http://www.springer.com/lncs}. Last accessed 4
% \end{thebibliography}
% will be separated into its own document before submitting
\appendix
\section{Generated anomaly maps from different model configurations}
\begin{figure}[ht]
    \centering
    \begin{subfigure}[t]{0.13\textwidth}
        \includegraphics[height=0.75in]{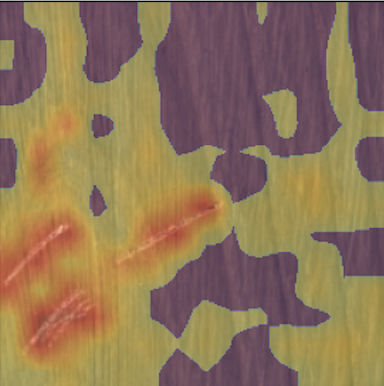}
    \end{subfigure}
    \hspace{0.5em}
    \begin{subfigure}[t]{0.13\textwidth}
        \includegraphics[height=0.75in]{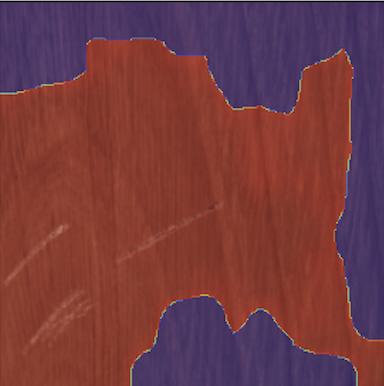}
    \end{subfigure}
    \hspace{0.5em}
    \begin{subfigure}[t]{0.13\textwidth}
        \includegraphics[height=0.75in]{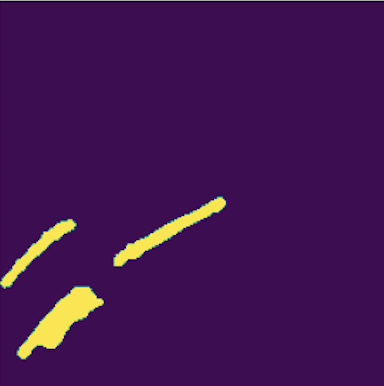}
    \end{subfigure}
    \hspace{0.5em}
    \begin{subfigure}[t]{0.13\textwidth}
        \includegraphics[height=0.75in]{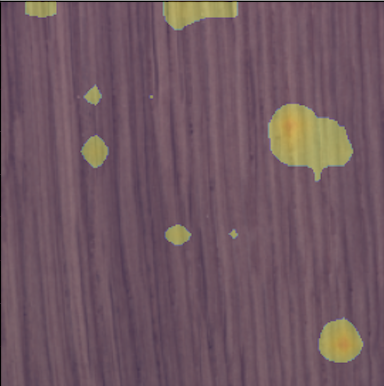}
    \end{subfigure}
    \hspace{0.5em}
    \begin{subfigure}[t]{0.13\textwidth}
        \includegraphics[height=0.75in]{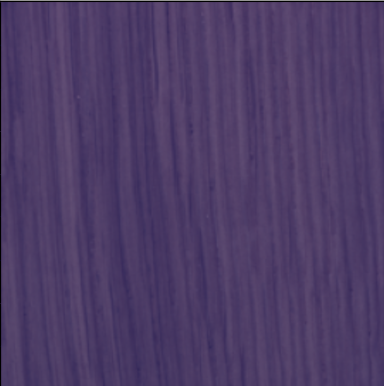}
    \end{subfigure}
    \hspace{0.5em}
    \begin{subfigure}[t]{0.13\textwidth}
        \includegraphics[height=0.75in]{images/gt_good.png}
    \end{subfigure}
    \begin{subfigure}[t]{0.13\textwidth}
        \includegraphics[height=0.75in]{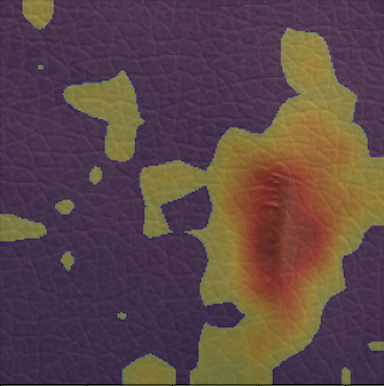}
    \end{subfigure}
    \hspace{0.5em}
    \begin{subfigure}[t]{0.13\textwidth}
        \includegraphics[height=0.75in]{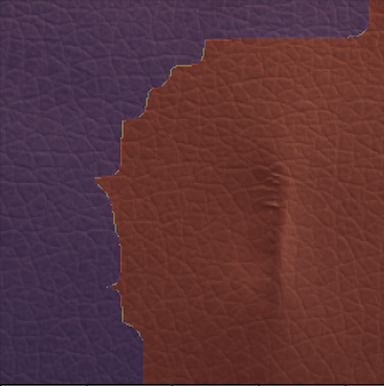}
    \end{subfigure}
    \hspace{0.5em}
    \begin{subfigure}[t]{0.13\textwidth}
        \includegraphics[height=0.75in]{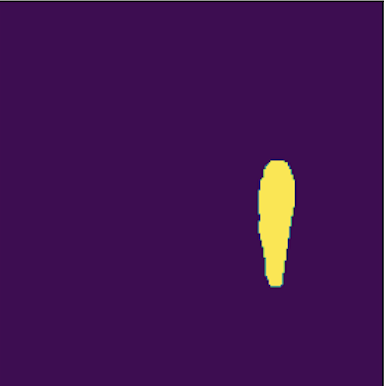}
    \end{subfigure}
    \hspace{0.5em}
    \begin{subfigure}[t]{0.13\textwidth}
        \includegraphics[height=0.75in]{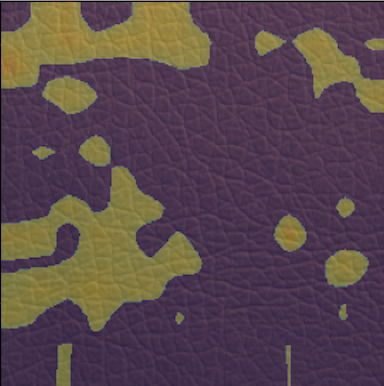}
    \end{subfigure}
    \hspace{0.5em}
    \begin{subfigure}[t]{0.13\textwidth}
        \includegraphics[height=0.75in]{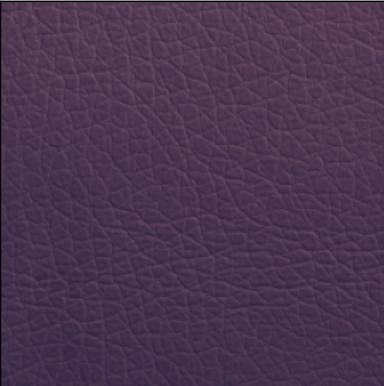}
    \end{subfigure}
    \hspace{0.5em}
    \begin{subfigure}[t]{0.13\textwidth}
        \includegraphics[height=0.75in]{images/gt_good.png}
    \end{subfigure}
    \begin{subfigure}[t]{0.13\textwidth}
        \includegraphics[height=0.75in]{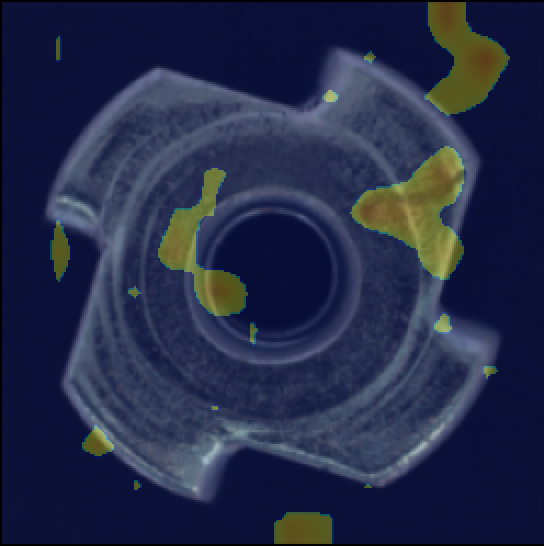}
    \end{subfigure}
    \hspace{0.5em}
    \begin{subfigure}[t]{0.13\textwidth}
        \includegraphics[height=0.75in]{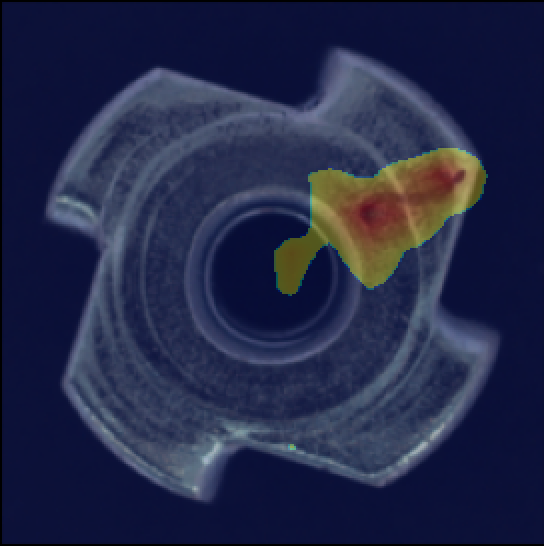}
    \end{subfigure}
    \hspace{0.5em}
    \begin{subfigure}[t]{0.13\textwidth}
        \includegraphics[height=0.75in]{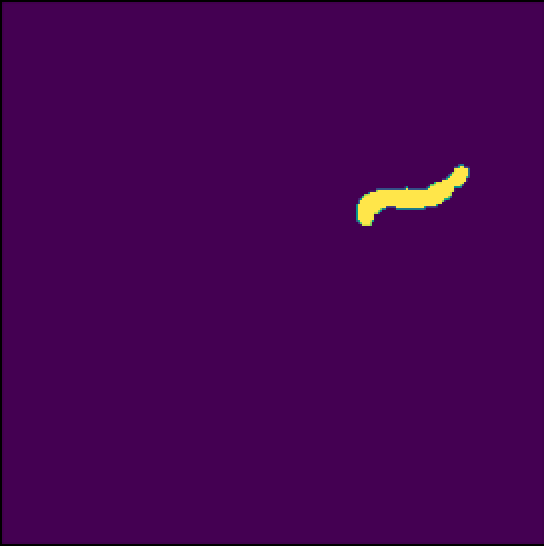}
    \end{subfigure}
    \hspace{0.5em}
    \begin{subfigure}[t]{0.13\textwidth}
        \includegraphics[height=0.75in]{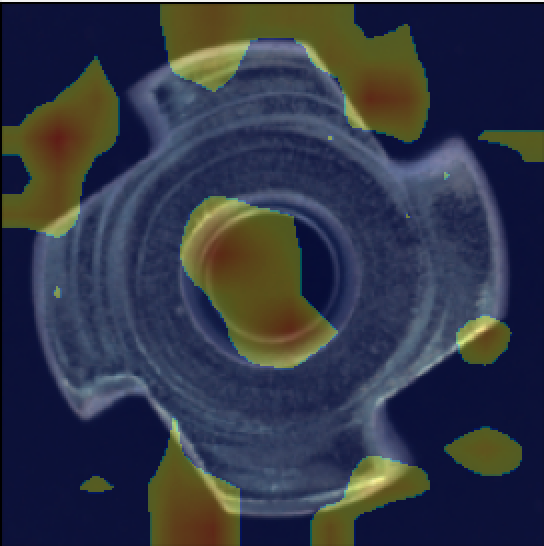}
    \end{subfigure}
    \hspace{0.5em}
    \begin{subfigure}[t]{0.13\textwidth}
        \includegraphics[height=0.75in]{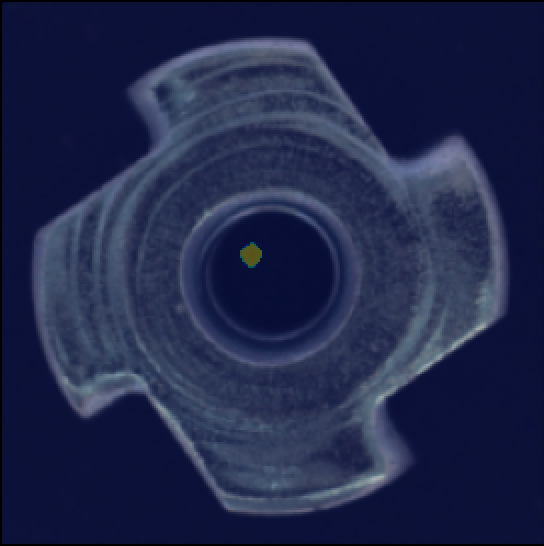}
    \end{subfigure}
    \hspace{0.5em}
    \begin{subfigure}[t]{0.13\textwidth}
        \includegraphics[height=0.75in]{images/gt_good.png}
    \end{subfigure}
    \begin{subfigure}[t]{0.13\textwidth}
        \includegraphics[height=0.75in]{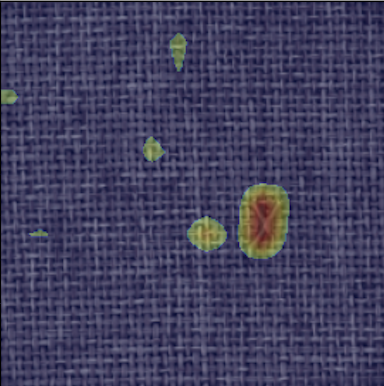}
    \end{subfigure}
    \hspace{0.5em}
    \begin{subfigure}[t]{0.13\textwidth}
        \includegraphics[height=0.75in]{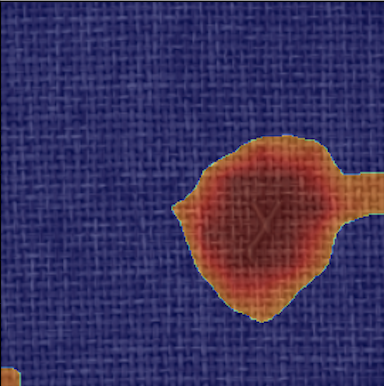}
    \end{subfigure}
    \hspace{0.5em}
    \begin{subfigure}[t]{0.13\textwidth}
        \includegraphics[height=0.75in]{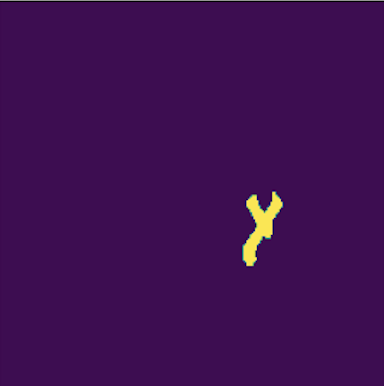}
    \end{subfigure}
    \hspace{0.5em}
    \begin{subfigure}[t]{0.13\textwidth}
        \includegraphics[height=0.75in]{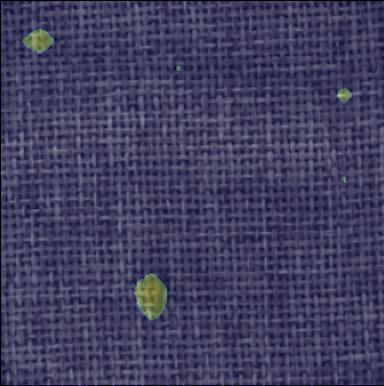}
    \end{subfigure}
    \hspace{0.5em}
    \begin{subfigure}[t]{0.13\textwidth}
        \includegraphics[height=0.75in]{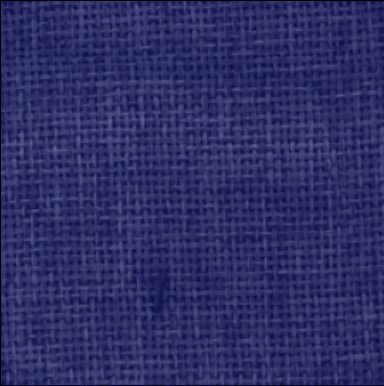}
    \end{subfigure}
    \hspace{0.5em}
    \begin{subfigure}[t]{0.13\textwidth}
        \includegraphics[height=0.75in]{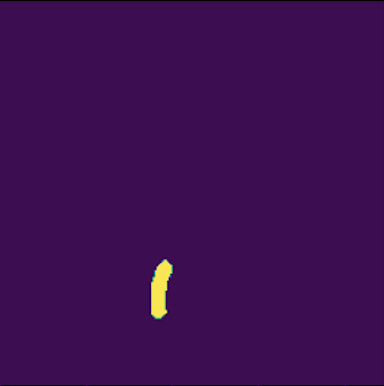}
    \end{subfigure}
    \begin{subfigure}[t]{0.13\textwidth}
        \includegraphics[height=0.75in]{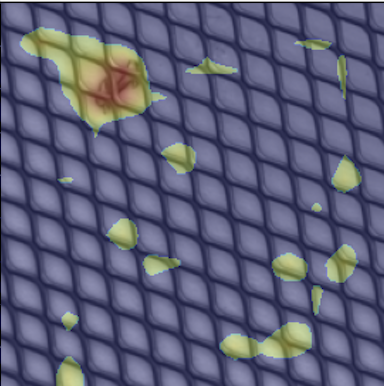}
    \end{subfigure}
    \hspace{0.5em}
    \begin{subfigure}[t]{0.13\textwidth}
        \includegraphics[height=0.75in]{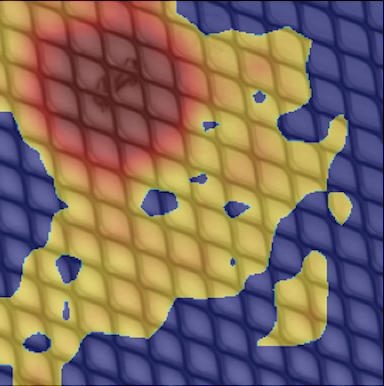}
    \end{subfigure}
    \hspace{0.5em}
    \begin{subfigure}[t]{0.13\textwidth}
        \includegraphics[height=0.75in]{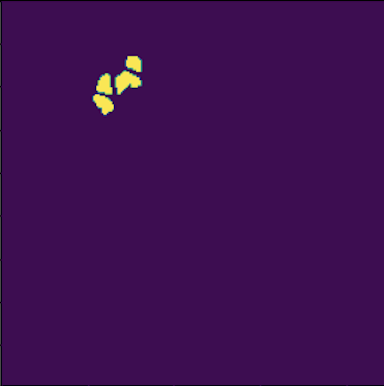}
    \end{subfigure}
    \hspace{0.5em}
    \begin{subfigure}[t]{0.13\textwidth}
        \includegraphics[height=0.75in]{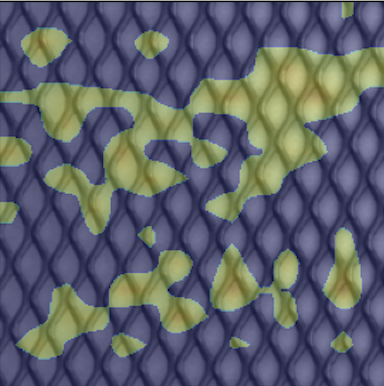}
    \end{subfigure}
    \hspace{0.5em}
    \begin{subfigure}[t]{0.13\textwidth}
        \includegraphics[height=0.75in]{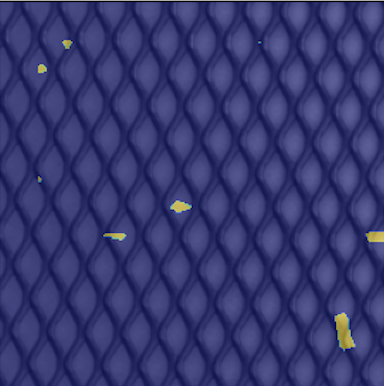}
    \end{subfigure}
    \hspace{0.5em}
    \begin{subfigure}[t]{0.13\textwidth}
        \includegraphics[height=0.75in]{images/gt_good.png}
    \end{subfigure}
    \caption{Anomaly maps for the classes \textit{wood}, \textit{leather}, \textit{metal nut}, \textit{carpet} and \textit{grid}, from the first row to the last, from the MVTecAD dataset with the NF model and the \textit{DeiT} encoder. In each row, the first anomaly map is created with NF i7, the second with NF i11 and the third image shows the corresponding ground truth. }\label{fig:comblocksnf}
\end{figure}

\begin{figure}[ht]
    \centering
    \begin{subfigure}[t]{0.15\textwidth}
        \includegraphics[height=0.85in]{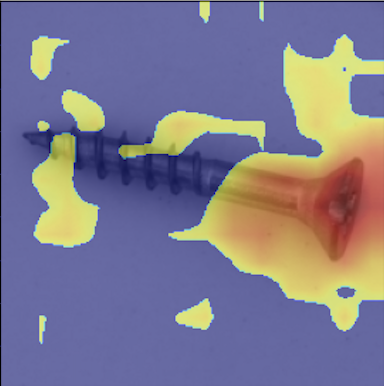}
    \end{subfigure}
    \hspace{0.5em}
    \begin{subfigure}[t]{0.15\textwidth}
        \includegraphics[height=0.85in]{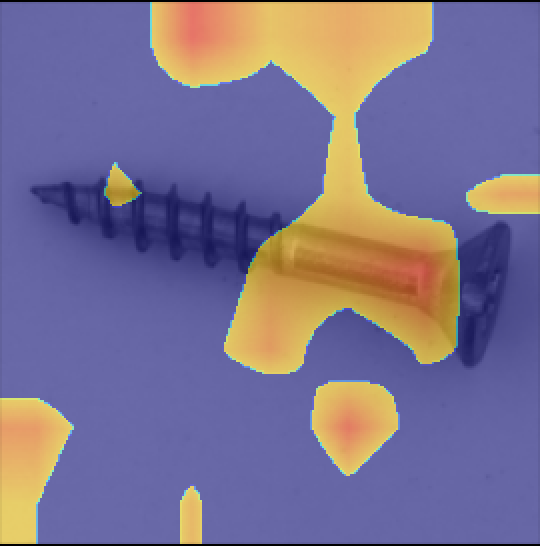}
    \end{subfigure}
    \hspace{0.5em}
    \begin{subfigure}[t]{0.15\textwidth}
        \includegraphics[height=0.85in]{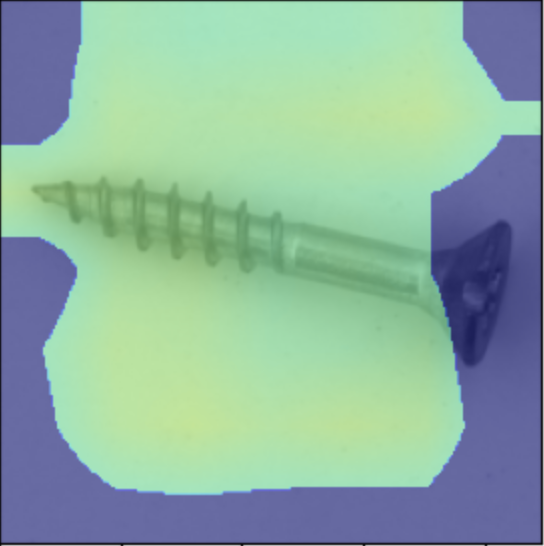}
    \end{subfigure}
    \hspace{0.5em}
    \begin{subfigure}[t]{0.15\textwidth}
        \includegraphics[height=0.85in]{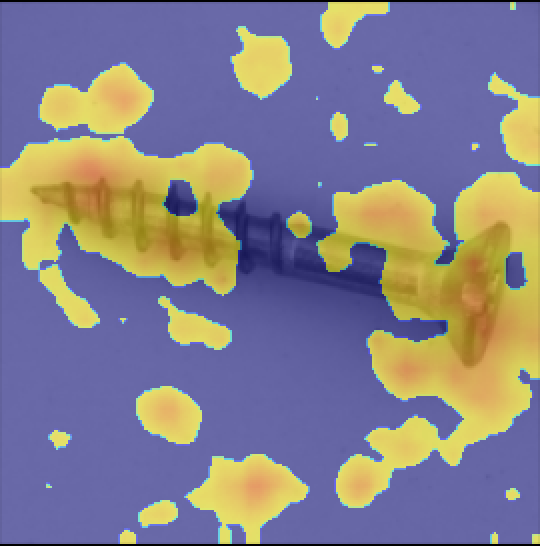}
    \end{subfigure}
    \hspace{0.5em}
    \begin{subfigure}[t]{0.15\textwidth}
        \includegraphics[height=0.85in]{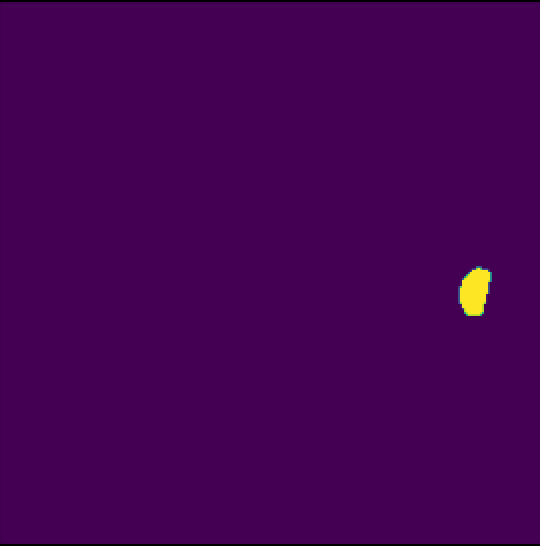}
    \end{subfigure}
    \begin{subfigure}[t]{0.15\textwidth}
        \includegraphics[height=0.85in]{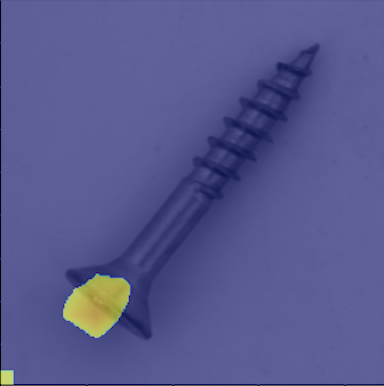}
    \end{subfigure}
    \hspace{0.5em}
    \begin{subfigure}[t]{0.15\textwidth}
        \includegraphics[height=0.85in]{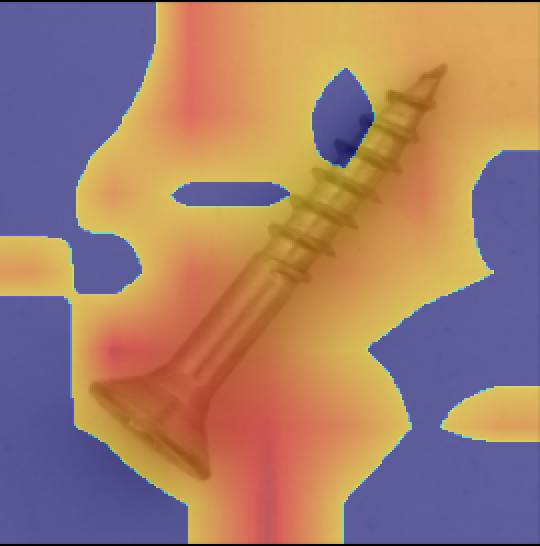}
    \end{subfigure}
    \hspace{0.5em}
    \begin{subfigure}[t]{0.15\textwidth}
        \includegraphics[height=0.85in]{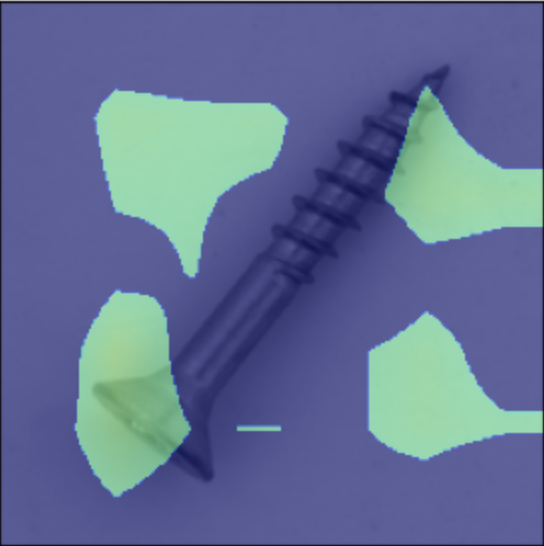}
    \end{subfigure}
    \hspace{0.5em}
    \begin{subfigure}[t]{0.15\textwidth}
        \includegraphics[height=0.85in]{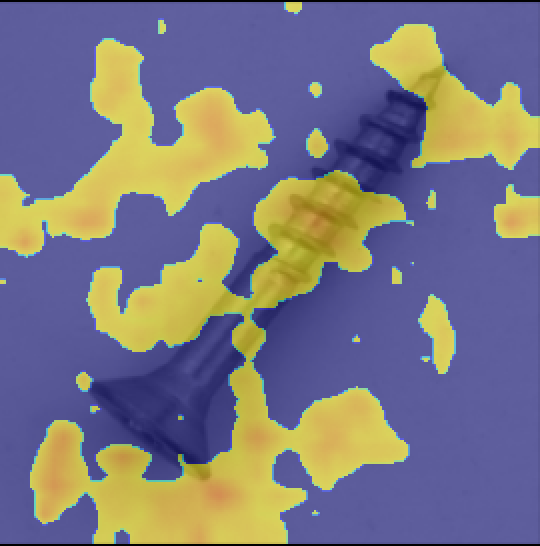}
    \end{subfigure}
    \hspace{0.5em}
    \begin{subfigure}[t]{0.15\textwidth}
        \includegraphics[height=0.85in]{images/gt_good.png}
    \end{subfigure}
    \caption{Anomaly maps for the \textit{screw} class of the NF model. The used backbones are from left to right: \textit{DeiT}, \textit{EfficientFormer}, \textit{EsViT}, \textit{ResNet} and the corresponding ground truth. It can be noted, that \textit{DeiT} was the only backbone with which the model was able to capture the small anomaly.}\label{fig:screwheatmapnf}
\end{figure}
\newpage

\section{Class distributions of the used datasets}
\begin{figure}
    \centering
    \includegraphics[height=2in]{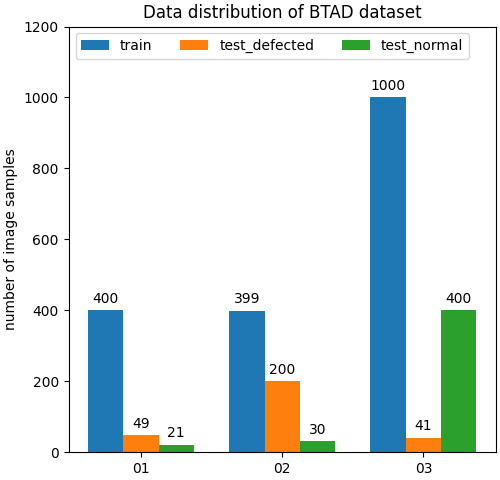}
    \caption{Distribution of the different classes of the BTAD dataset.}
    \label{fig:btaddist}
\end{figure}

\begin{figure}
    \centering
    \includegraphics[width=1\textwidth]{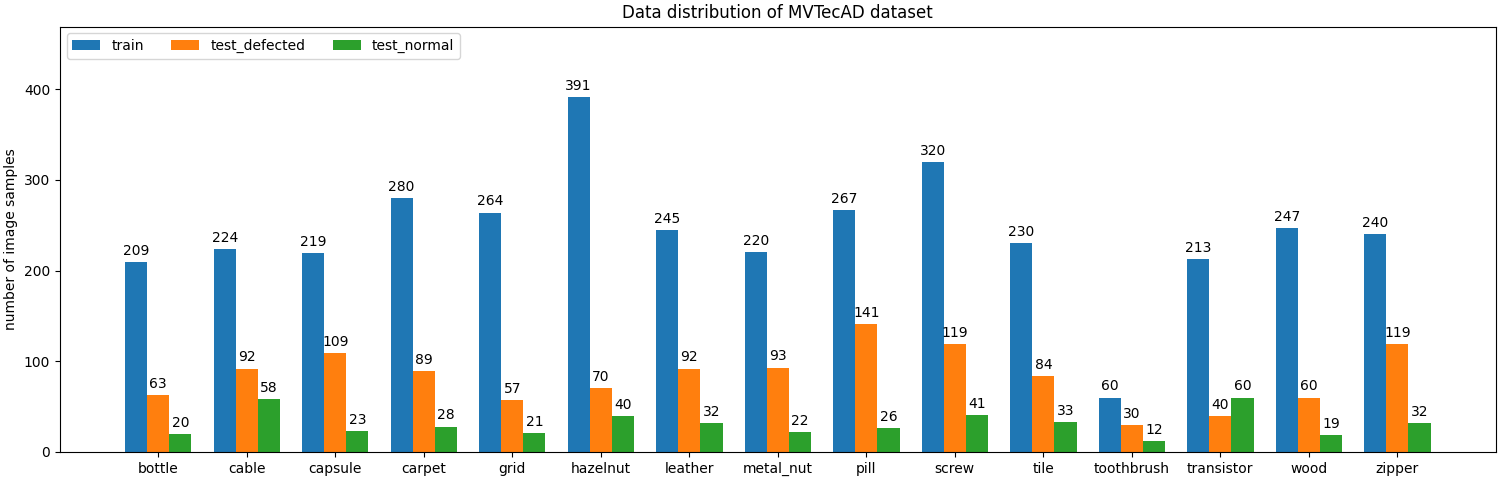}
    \caption{Distribution of the different classes of the MVTecAD dataset.}
    \label{fig:mvtecaddist}
\end{figure}
\newpage

\section{Ablation study and production application}
\begin{figure}[ht]
    \centering
    \begin{subfigure}[t]{0.49\textwidth}
        \includegraphics[height=1.7in]{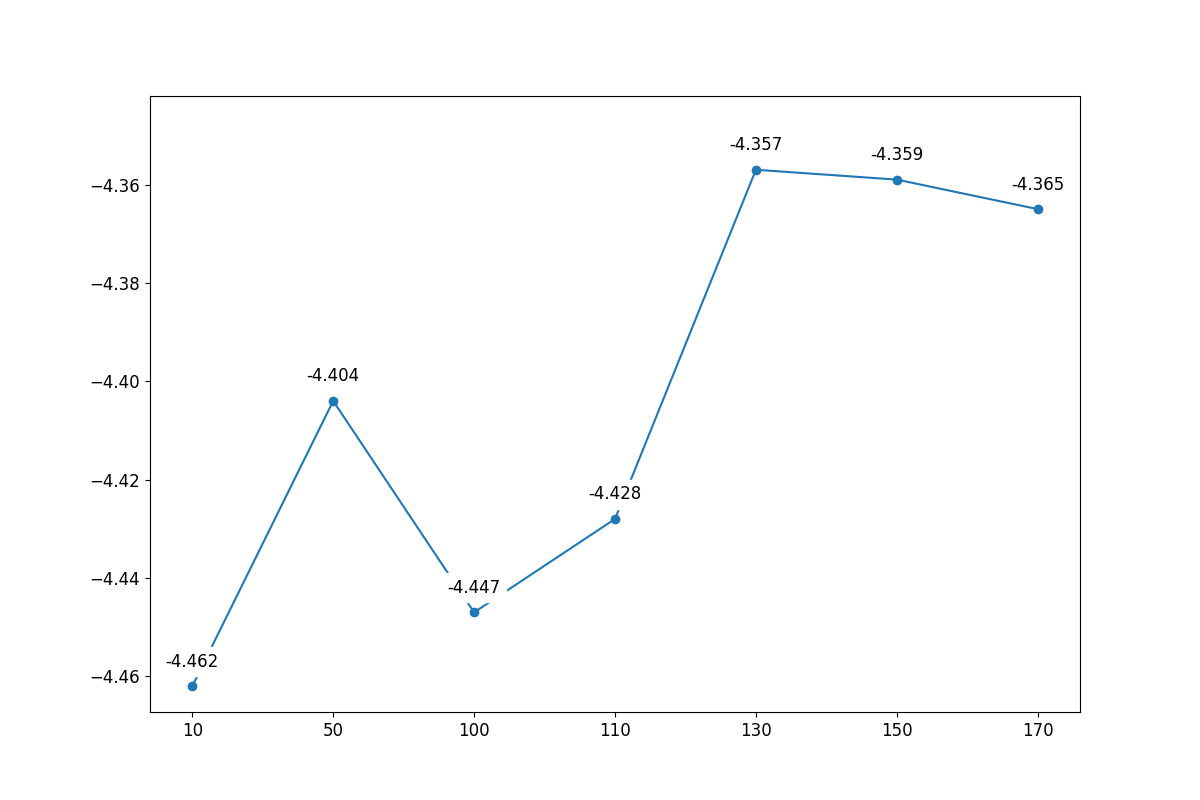}
    \end{subfigure}
    \begin{subfigure}[t]{0.49\textwidth}
        \includegraphics[height=1.7in]{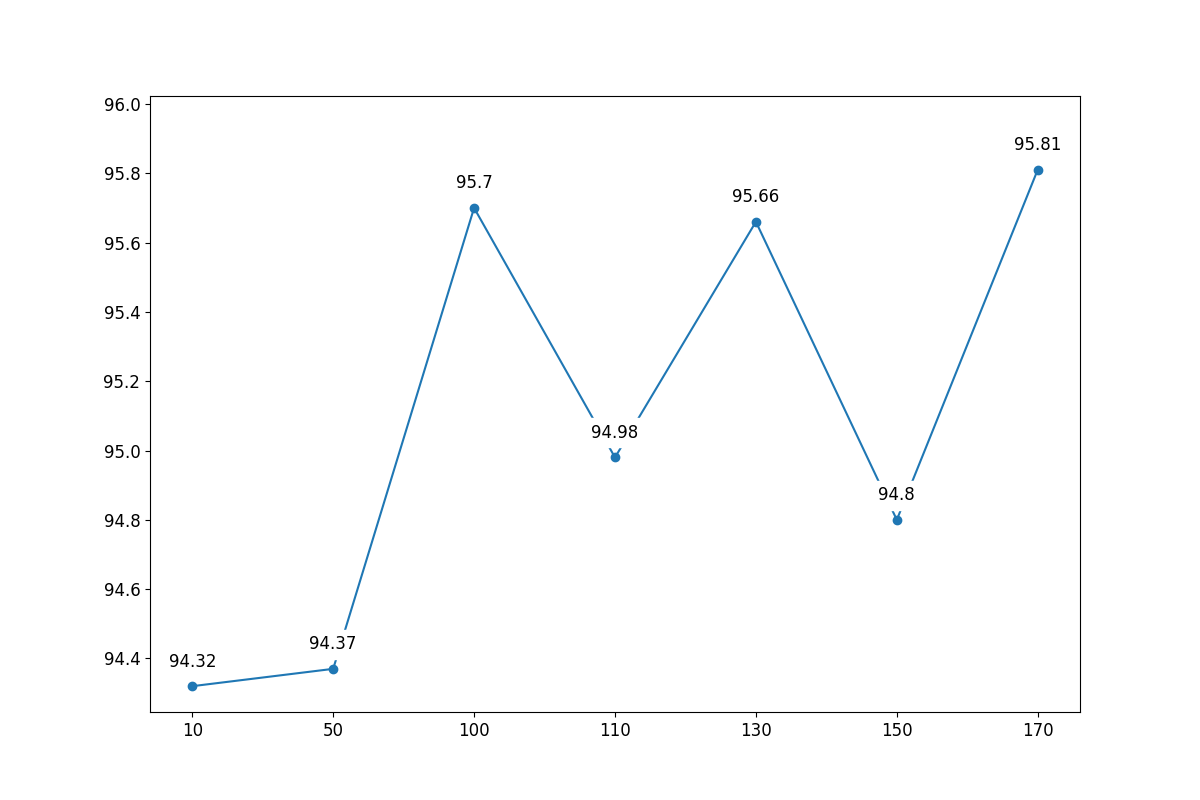}
    \end{subfigure}
    \caption{Plots of loss (left) and PRO score (right) for different numbers of Gaussian's on the data class \textit{hazelnut} from MVTecAD. The graphics show that there is no direct relation between PRO score and likelihood loss.}\label{fig:lossvspro}
\end{figure}

\begin{figure}
    \centering
    \includegraphics[width=1\textwidth]{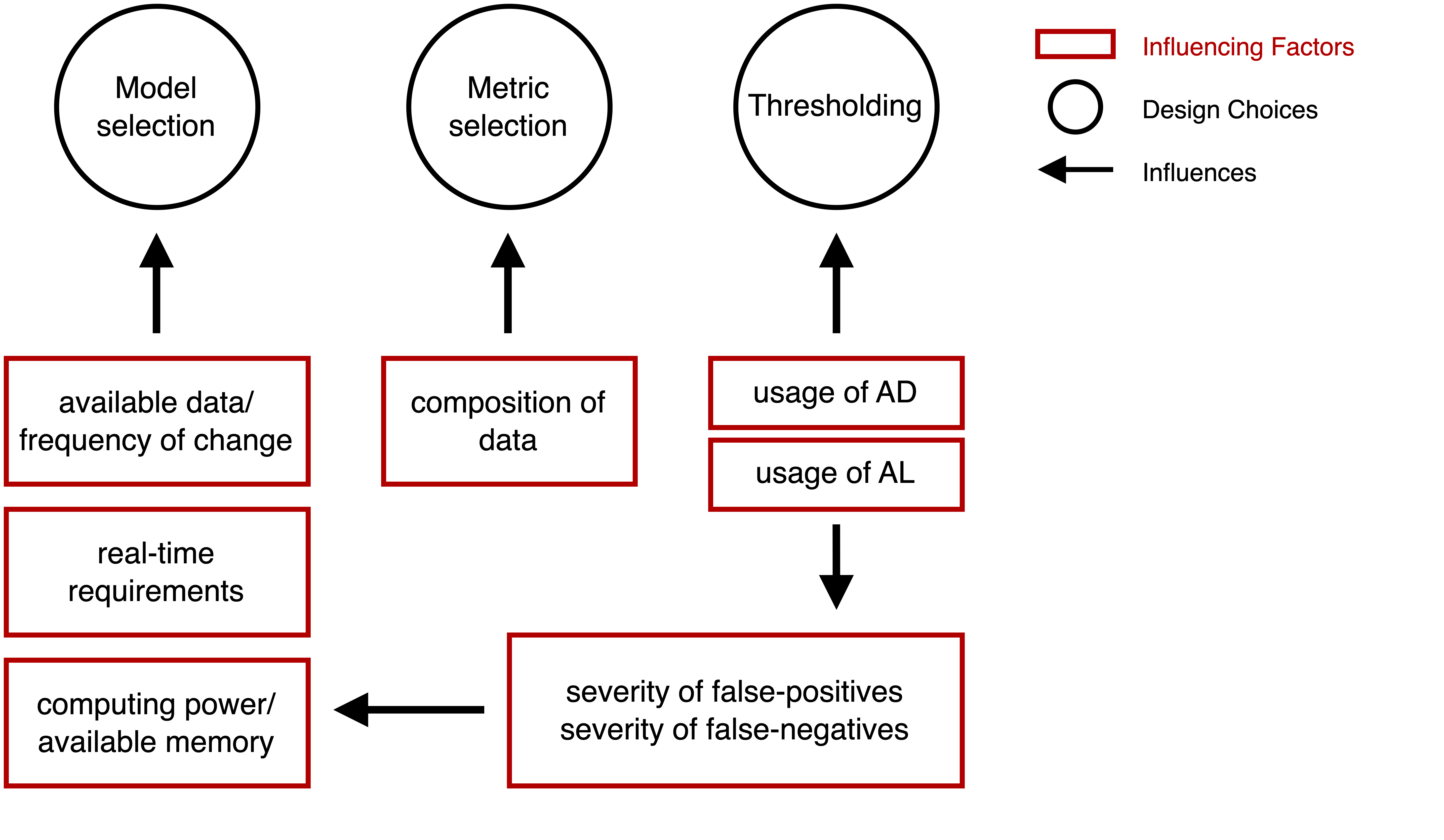}
    \caption{High level overview on decisions to be made when applying our approaches to production scenarios. }\label{fig:proddecision}
\end{figure}
\end{document}